\documentclass[preprint,12pt]{elsarticle}

\usepackage{amssymb}
\usepackage{amsmath}
\usepackage{graphicx}
\usepackage{subcaption}
\usepackage{float}
\usepackage{booktabs}
\usepackage{tabularx}
\usepackage{xcolor}
\usepackage{tikz}
\usepackage{url}
\usepackage[colorlinks=true,
            linkcolor=blue,
            citecolor=blue,
            urlcolor=blue]{hyperref}

\journal{Safety Science}

\begin{document}

\begin{frontmatter}

\title{Comparing Human Gaze and Vision-Language Model Attention in Safety-Relevant Environments}

\author[hwu]{Siwen Wang}
\author[hwu]{Marta Vallejo\corref{cor1}}
\ead{m.vallejo@hw.ac.uk}
\cortext[cor1]{Corresponding author}

\affiliation[hwu]{organization={School of Mathematical and Computer Sciences, Heriot-Watt University},
            addressline={Edinburgh},
            postcode={EH14 4AS},
            country={United Kingdom}}

\begin{abstract}
Human visual attention plays an important role in how people perceive and respond to environments containing potential risks. This study investigates whether large vision-language models can identify the same regions of a scene that attract human attention in safety-relevant environments. Eye-tracking data were collected from ten participants viewing 33 scene images representing environments with varying levels of potential risk using Pupil Invisible wearable glasses. Gaze coordinates were mapped onto stimulus images to generate population-averaged human gaze heatmaps. In parallel, GPT-4o was prompted through the OpenAI Vision Application Programming Interface (API) to generate spatial predictions of visual attention, which were converted into saliency maps for comparison with human gaze patterns.

Spatial alignment between human gaze heatmaps and model-generated saliency maps was evaluated using four complementary metrics: Pearson correlation ($r = 0.515 \pm 0.117$), Normalised Scanpath Saliency (NSS $= 0.988 \pm 0.323$), Kullback--Leibler divergence (KL $= 1.766 \pm 0.844$), and Area Under the Receiver Operating Characteristic Curve using the Judd formulation (AUC-Judd $= 0.806 \pm 0.076$). A cross-model comparison with Gemini Pro, Gemini Flash, and Claude showed that all models exceeded the AUC-Judd chance baseline of 0.5 and achieved positive NSS scores. Gemini Pro demonstrated the strongest spatial localisation according to three of the four metrics, whereas GPT-4o produced the closest distributional match to human attention as measured by KL divergence.

These findings suggest that large vision-language models can identify regions that broadly correspond to where humans direct visual attention in safety-relevant scenes without requiring eye-tracking training data. The results highlight the potential of vision-language models as a scalable tool for approximating human attentional patterns.
\end{abstract}


\end{frontmatter}

\section{Introduction}
\label{sec:intro}
Safety-aware artificial intelligence (AI) systems are increasingly being developed for applications in autonomous driving, robotics, and public space monitoring. A critical requirement for such systems is the ability to identify the same regions of a scene that humans perceive as potentially risky. Understanding whether AI models can replicate human spatial attention in safety-relevant scenes therefore has both theoretical and practical significance. Eye-tracking technology provides a direct measure of visual attention, revealing how individuals allocate gaze across complex visual environments and safety-relevant scenes \cite{Ahlstrom2021,Kang2026}.

A central insight from psychological research is that human attention is not neutral: people are disproportionately sensitive to negative or threatening stimuli, a phenomenon known as negativity bias \cite{Baumeister2001Bad,Rozin2001Negativity}. This means that when viewing a scene, humans tend to fixate more on elements that represent potential danger. Ohman demonstrated this concretely, showing that threat-relevant targets were detected faster and more efficiently than neutral ones, suggesting that risk-related processing operates in a near-automatic, pre-attentive manner \cite{Ohman2001Snake}. Together, these findings imply that human gaze heatmaps in safety-relevant scenes should reflect a meaningful spatial concentration of attention on regions containing potential risks.

This raises a corresponding question about AI: can a large vision-language model (VLM), prompted to identify potential risks in a scene, produce spatial predictions that align with where humans actually look? Recent advances in multimodal models such as GPT-4o have enabled structured, spatially grounded risk predictions from scene images without requiring human participants \cite{OpenAI2024HelloGPT4o}. Recent advances in computational saliency modelling have demonstrated that deep learning systems can predict human gaze allocation with increasing accuracy across natural scenes \cite{Kummerer2022}. These approaches have shown that learned visual representations can capture important aspects of human attentional behaviour, providing a foundation for comparing model-generated attention maps with human eye-tracking data. However, most existing work has focused on predicting visual saliency rather than safety perception, and has not examined whether attention maps generated by modern vision-language models align spatially with human gaze in safety-relevant environments. Whether such alignment can be quantified using established saliency metrics remains an open question.

Existing eye-tracking research on safety perception has often focused on specific domains such as construction safety and hazard recognition \cite{Hasanzadeh2016,Jeelani2018}. Similarly, substantial eye-tracking datasets have been developed for attention-based tasks in autonomous and assisted driving \cite{Ahlstrom2021,Alletto2016DReyeVE}. However, these studies and datasets are typically tied to a single application domain, whereas cross-environment datasets spanning diverse everyday scene types remain limited. A dataset spanning multiple environment types is necessary to determine whether AI--human spatial alignment is consistent across contexts, or whether it varies depending on scene content and layout. The present study therefore constructs a dedicated eye-tracking dataset covering a range of everyday scene types with stimuli varying in the level of potential risk. This dataset provides the population-averaged human gaze heatmaps that serve as the human ground truth for evaluating AI spatial predictions.

\begin{figure}[H]
\centering
\includegraphics[width=0.6\textwidth]{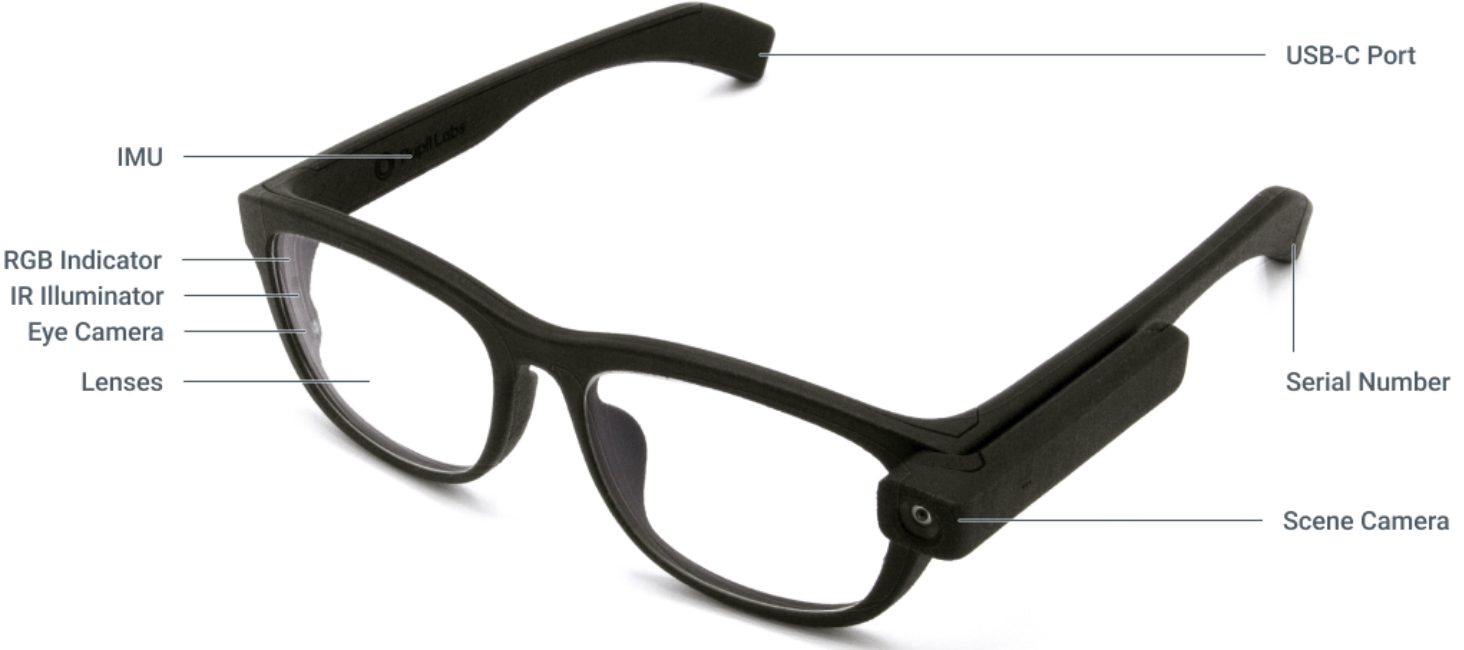}
\caption{Pupil Invisible hardware structure, including dual infrared eye cameras and a forward-facing scene camera for naturalistic gaze recording.}
\label{fig:hardware}
\end{figure}

To record gaze data, the study uses the Pupil Invisible eye-tracking glasses (see Figure~\ref{fig:hardware}). The device integrates two infrared eye cameras for binocular gaze tracking and a forward-facing scene camera that records the visual field. The glasses are designed for unobtrusive eye-tracking in natural viewing conditions and allow participants to view images comfortably while providing measurements of gaze behaviour and pupil size. These characteristics make the device well suited for studying visual attention across diverse scene types \cite{PupilLabs2025,Kassner2014}.

The study pursues two objectives. The first is to construct a curated eye-tracking dataset spanning various everyday scene categories using the Pupil Invisible wearable glasses, covering 33 stimulus images with varying levels of potential risk, and generating population-averaged fixation density maps for each stimulus. The second is to develop an automated Python pipeline to generate spatially grounded risk heatmaps for each stimulus using GPT-4o through the OpenAI Vision API, without any training on eye-tracking data, and to evaluate their spatial alignment with the population-averaged human gaze heatmaps using Pearson correlation, Normalised Scanpath Saliency (NSS), Kullback--Leibler (KL) divergence, and AUC-Judd.

\section{Background}
\label{sec:background}

\subsection{Visual Attention and Perception of Potential Risks}
Eye-tracking technology provides a direct measure of visual attention by recording gaze coordinates over time. Fixation duration and spatial distribution offer insights into how visual attention is allocated within a scene and have been widely used to investigate attentional processes in domains such as driving, hazard perception, and human–environment interaction \cite{Ahlstrom2021,Kang2026}. Patterns of gaze behaviour can therefore help identify which elements of a scene attract attention and how individuals respond to environments that may contain potential risks. 

These gaze patterns are not uniformly distributed: a key insight from psychological research is that human attention is disproportionately sensitive to negative or threatening stimuli, a phenomenon known as negativity bias \cite{Baumeister2001Bad,Rozin2001Negativity}. Ohman demonstrated this concretely in a visual search paradigm, showing that threat-relevant targets were detected faster and more efficiently than neutral ones, suggesting that risk-related processing operates in a near-automatic, pre-attentive manner \cite{Ohman2001Snake}. Consistent with this, eye-tracking studies of hazard recognition have shown that attention is preferentially allocated to safety-relevant regions within a scene and that gaze behaviour is associated with the successful detection of hazards \cite{Hasanzadeh2016}. Together, these findings imply that human gaze heatmaps in safety-relevant scenes should reflect a meaningful spatial concentration of attention on regions containing potential risks, making them a theoretically grounded reference for evaluating AI spatial predictions.

\subsection{Eye-Tracking Datasets for Safety Perception}
Several eye-tracking datasets and experimental studies have been developed to investigate visual attention in safety-relevant contexts. In construction safety research, eye tracking has been used to examine hazard recognition, attentional allocation, and the influence of safety knowledge on visual search behaviour in both real and simulated environments \cite{Hasanzadeh2016,Jeelani2018,Hasanzadeh2018}. In driving research, datasets such as DR(eye)VE provide large-scale gaze recordings collected during real-world driving tasks and have supported the development and evaluation of attention-aware systems for autonomous and assisted driving \cite{Alletto2016DReyeVE}. More recently, Kang et al.\ collected eye-tracking data from participants viewing urban street-view images and compared population-level gaze heatmaps with attention maps generated by an explainable AI model \cite{Kang2026}, demonstrating that AI-derived attention maps can partially capture where humans look in outdoor urban environments.

However, existing datasets and studies remain largely confined to a single application domain, such as construction sites, driving scenarios, or urban street scenes. Consequently, comparatively little work has examined visual attention across multiple environment types using a common experimental framework, making it difficult to determine whether patterns of attention towards potential risks remain consistent across contexts. The dataset constructed in this study addresses this limitation by spanning a range of everyday scene types with systematically varying levels of potential risk, providing a broader empirical basis for evaluating AI--human spatial alignment across environments.

\subsection{AI-Based Approaches for Gaze and Risk Prediction}
Before the emergence of large VLMs, computational approaches to predicting human gaze relied primarily on bottom-up saliency models based on low-level visual features such as colour contrast, intensity, and spatial frequency \cite{Kummerer2018SaliencyBenchmarking,Bylinskii2019SaliencyMetrics}. While these models are computationally efficient, they lack semantic understanding and cannot reason about scene-specific risk content. Recent advances in large VLMs offer a new approach: models such as GPT-4o demonstrate strong capabilities in visual understanding and spatially grounded reasoning about scene content \cite{OpenAI2024HelloGPT4o}, and can be explicitly prompted to reason about potential risks in a scene.

An important question is whether AI models can exhibit human-like attentional biases towards elements that represent potential risks. Recent advances in computational saliency modelling have shown that deep learning systems can predict human gaze allocation with increasing accuracy across natural scenes, demonstrating that learned visual representations can capture important aspects of human attentional behaviour \cite{Kummerer2022}. However, most existing work has focused on visual saliency prediction rather than safety perception and has not examined whether attention maps generated by modern vision-language models align spatially with human gaze in safety-relevant environments. More systematic evaluation across diverse scene types remains limited, and consistent frameworks for comparing AI spatial predictions with human gaze behaviour remain underdeveloped. Saliency evaluation metrics such as AUC-Judd, Normalised Scanpath Saliency, and Pearson correlation provide a principled basis for this comparison \cite{Kummerer2018SaliencyBenchmarking,Bylinskii2019SaliencyMetrics}.

The present study addresses this gap by directly evaluating whether GPT-4o, prompted to identify potential risks, produces spatial predictions that align with population-averaged human gaze heatmaps across 33 diverse scene images.

\section{Materials and Methods}
\label{sec:methods}

\subsection{Experimental Setup}

\subsubsection{Stimuli}
Thirty-three scene images were used as visual stimuli, covering streets, indoor corridors, urban public spaces and green open areas. Some images contain potential risks such as vehicles, pedestrian crossings or obstacles, while others represent safe and neutral environments. Each image was shown for 10 seconds, followed by a 2-second blank interval before the next image.

All images were standardised to a resolution of $1{,}536 \times 1{,}024$ pixels and displayed with consistent brightness. Stimuli were presented on the same monitor for all participants. Display settings were kept identical throughout the study, including brightness, contrast, and colour profile. Each image was displayed for 10 seconds on a 24-inch monitor positioned 60-70 cm from the participant. Sessions were conducted in the same indoor room under stable artificial lighting, with blinds/curtains kept closed so ambient illumination did not vary between participants. Appointments were scheduled at a consistent time of day across participants.

\begin{figure}[H]
\centering
\begin{tabular}{ccc}
  \includegraphics[width=0.3\textwidth]{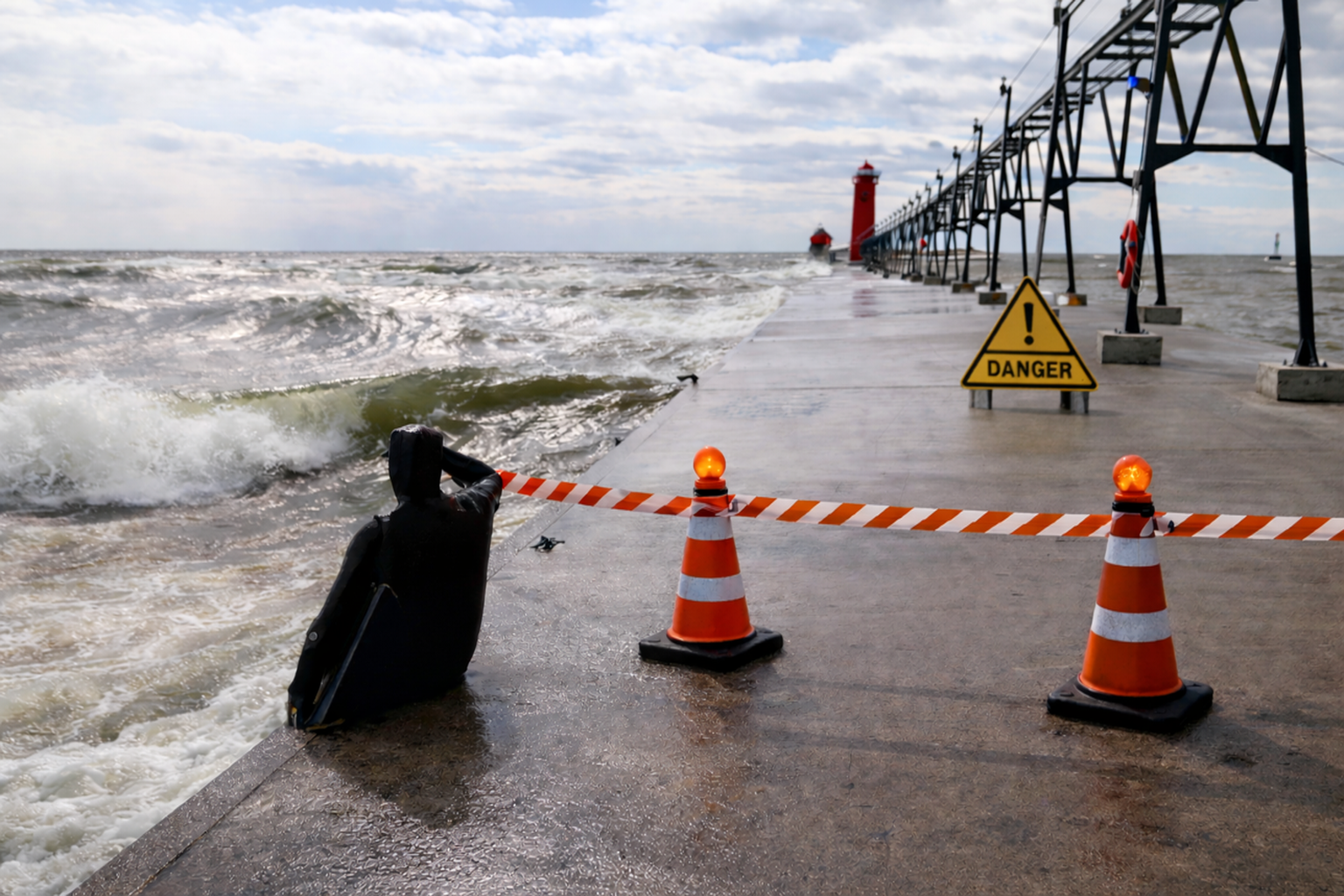} &
  \includegraphics[width=0.3\textwidth]{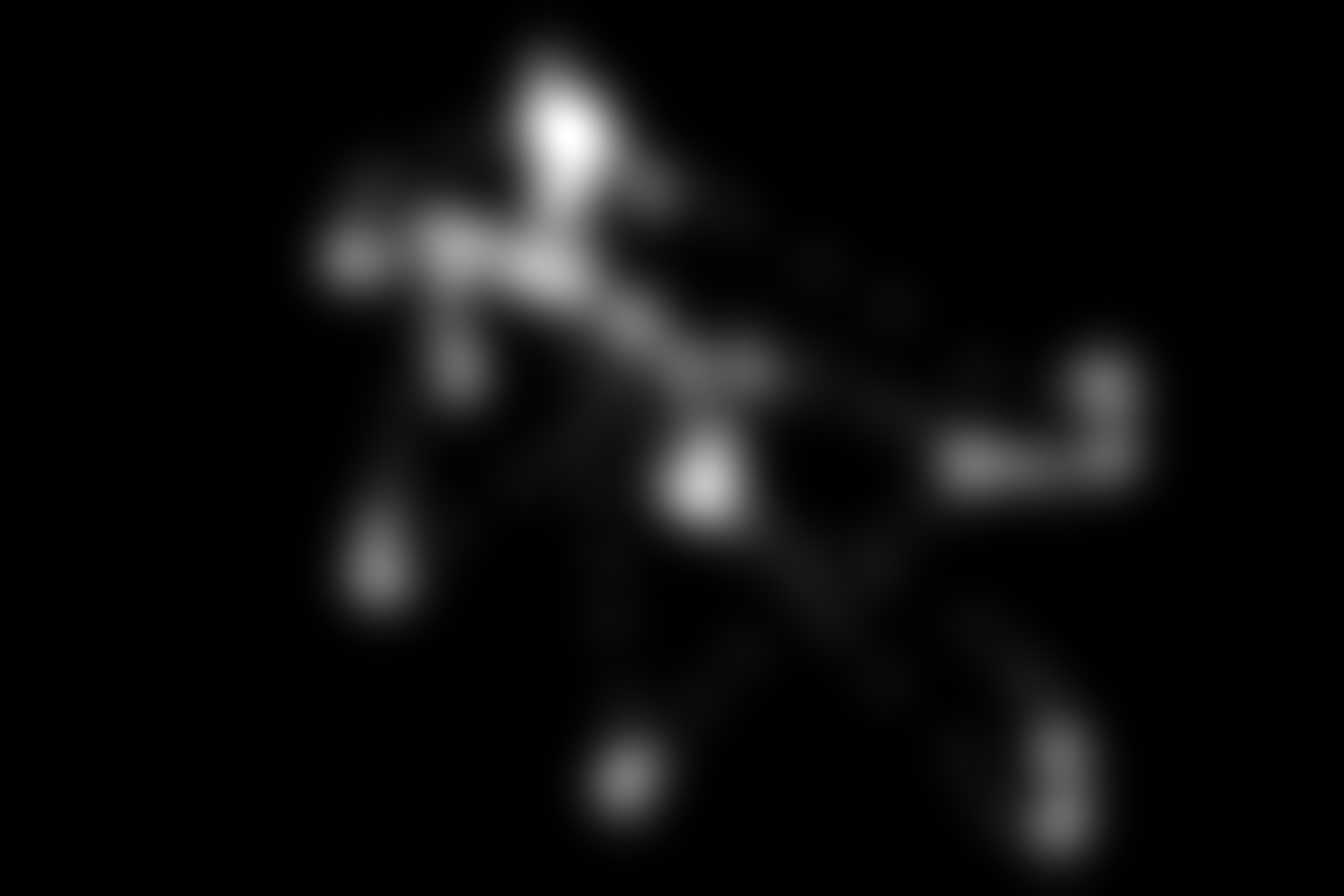} &
  \includegraphics[width=0.3\textwidth]{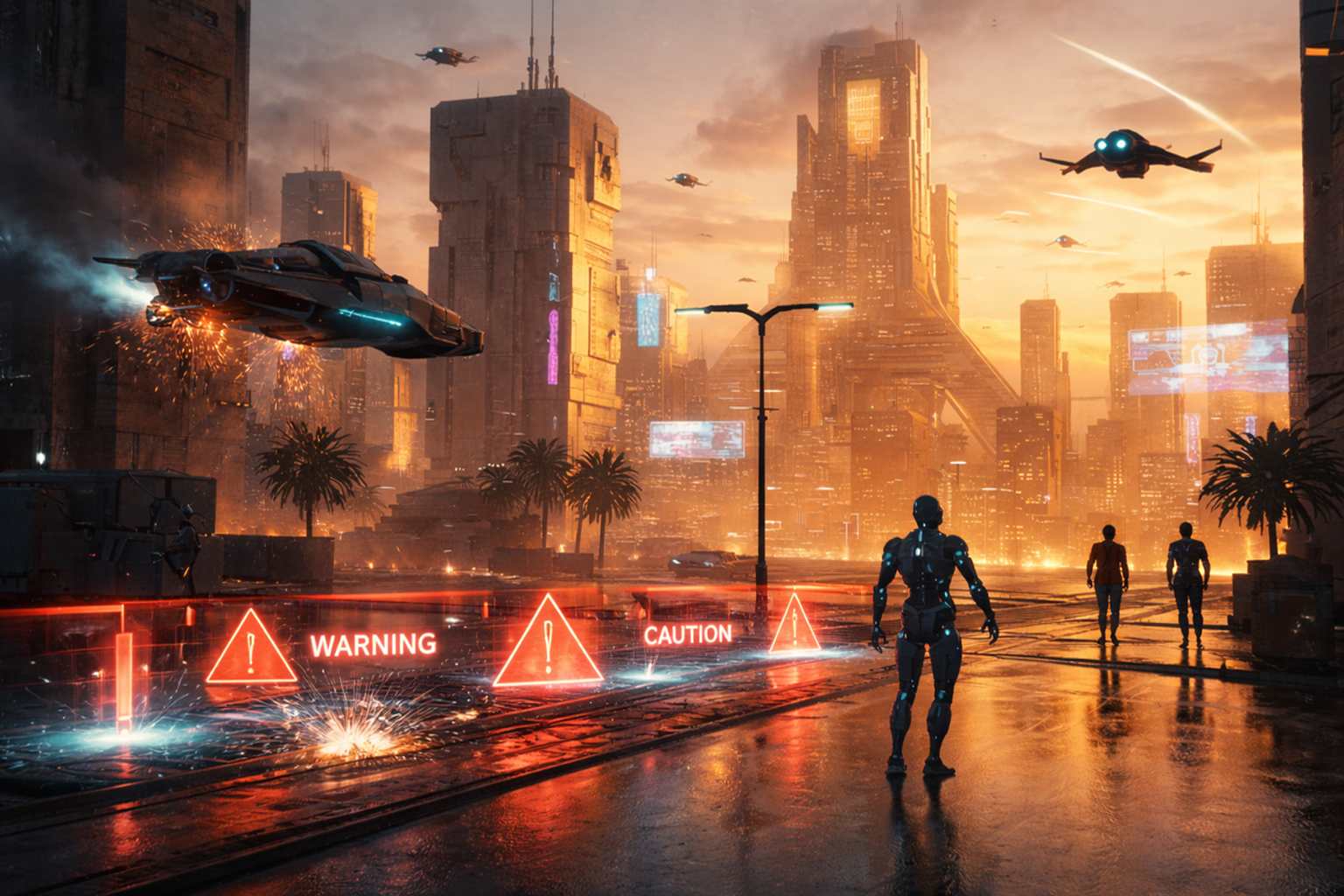} \\
  (a) Residential flood & (b) Coastal storm & (c) Urban flood \\[6pt]
  \includegraphics[width=0.3\textwidth]{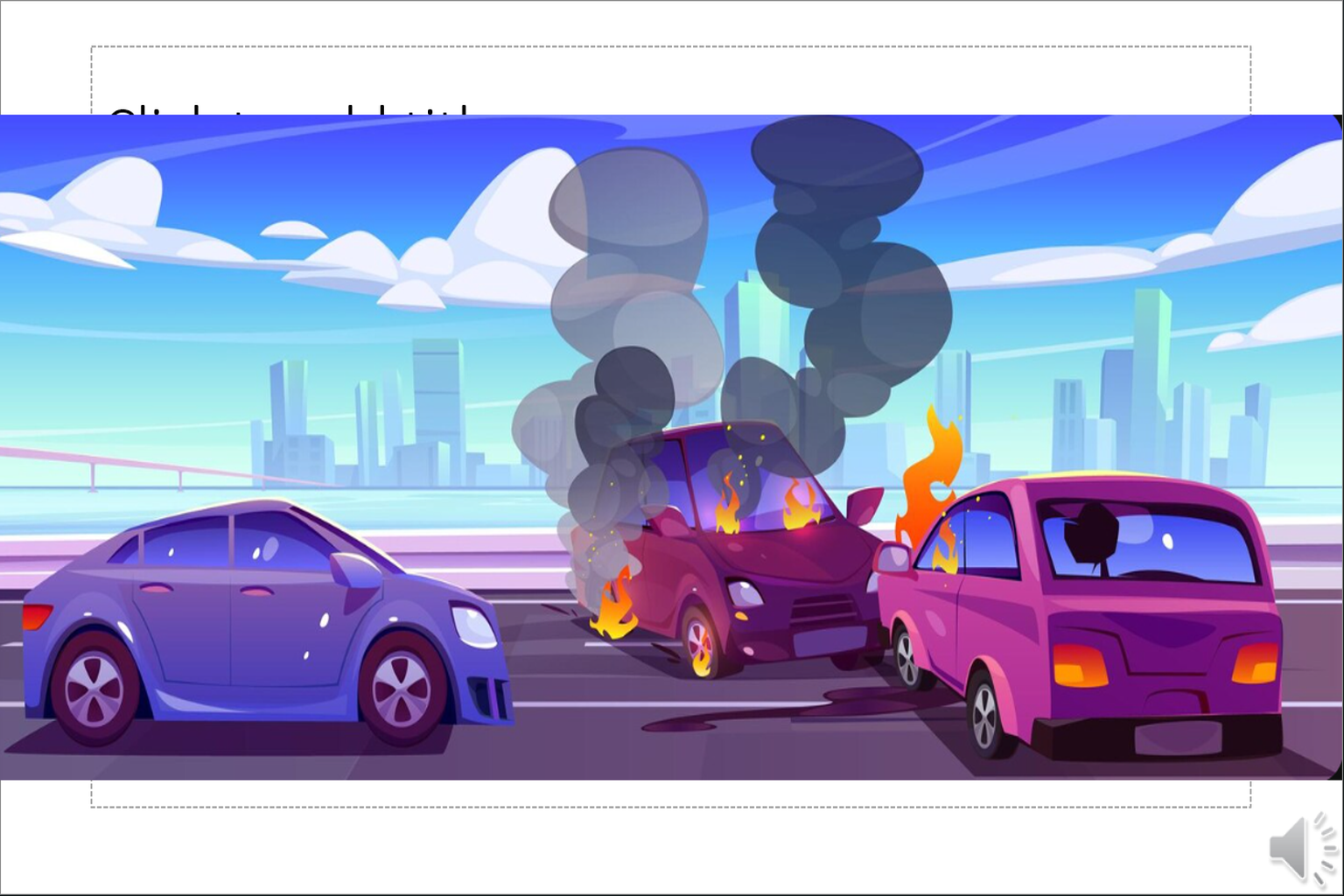} &
  \includegraphics[width=0.3\textwidth]{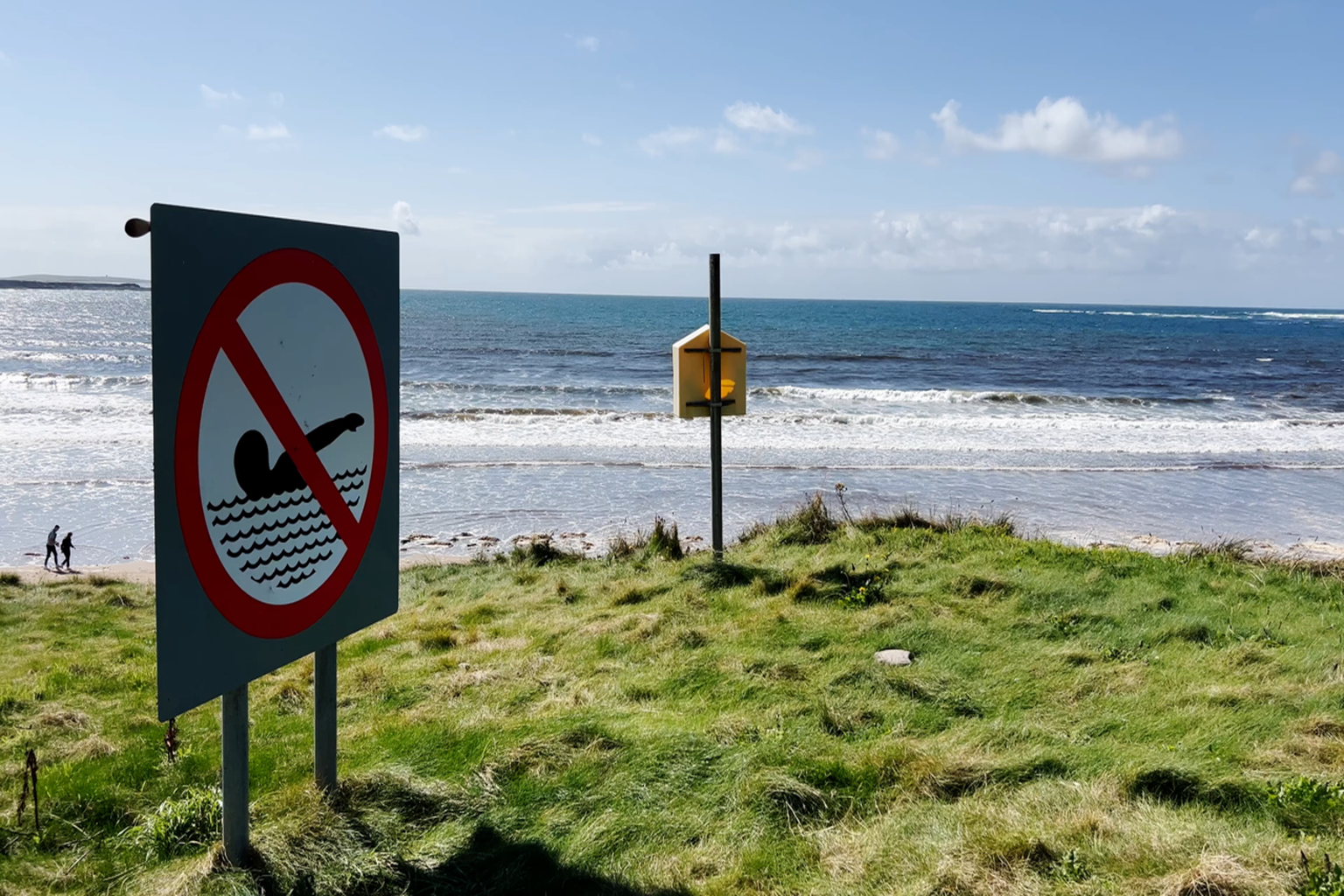} &
  \includegraphics[width=0.3\textwidth]{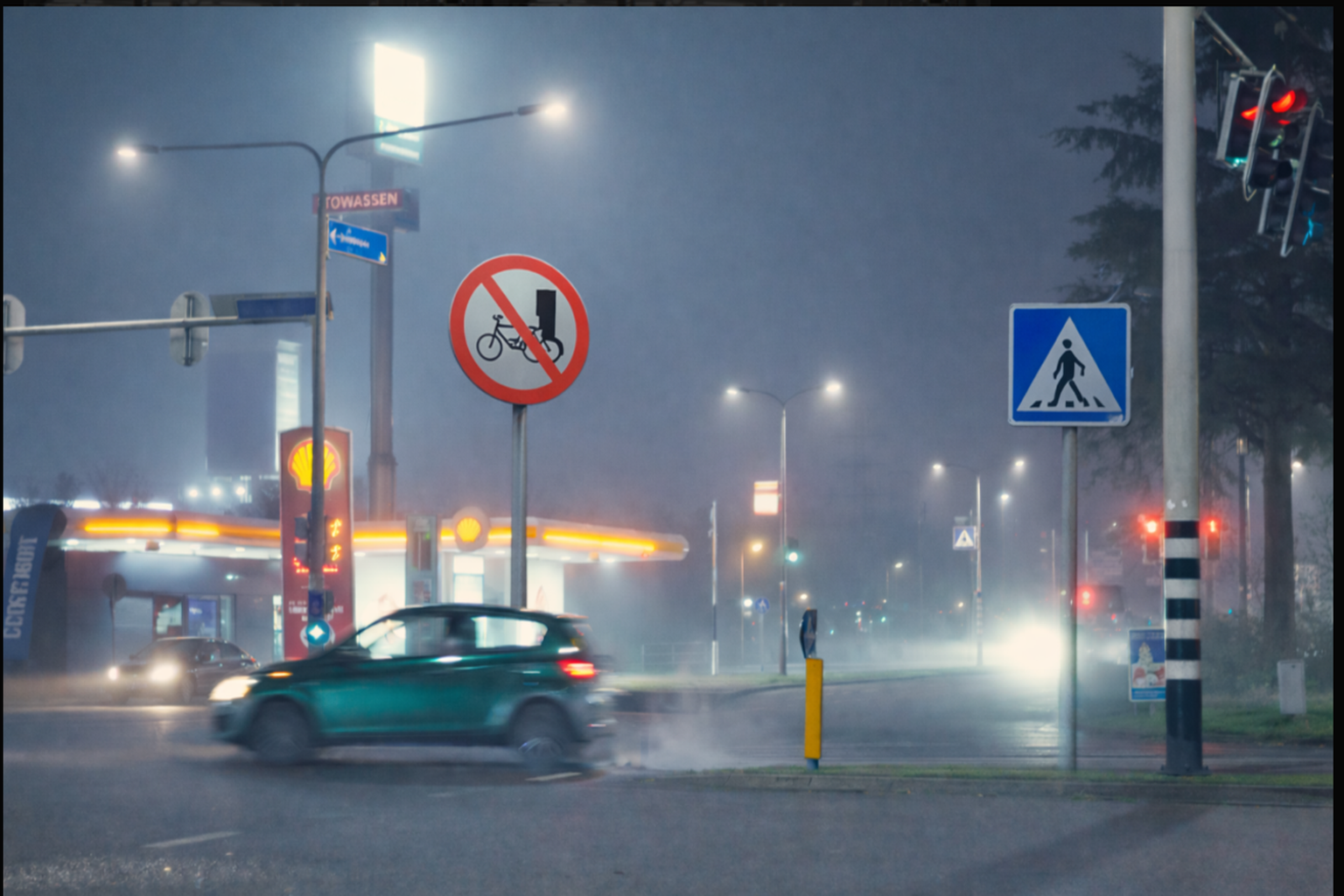} \\
  (d) Sci-fi risks & (e) Traffic accident & (f) Coastal signage \\
\end{tabular}
\caption{Examples of stimulus images from the 33-image set. Source: Freepik (\url{www.freepik.com}).}
\label{fig:stimuli_examples}
\end{figure}

\subsubsection{Participants}

Ten university students (6 female, 4 male) participated in the study. All had normal or corrected-to-normal vision and provided written informed consent prior to participation. Before the experiment, participants were informed that they would view a series of scene images, each displayed for 10 seconds. To encourage natural viewing behaviour, participants were told that they would later be asked to select the images they liked most and least. This incidental task ensured that attention was not explicitly directed towards risk detection. All participants provided written informed consent prior to participation. Participants were free to withdraw from the study at any time without providing a reason. The study was approved by the Mathematical and Computer Sciences Ethics Committee at Heriot-Watt University.

\begin{figure}[H]
\centering
\includegraphics[width=\textwidth]{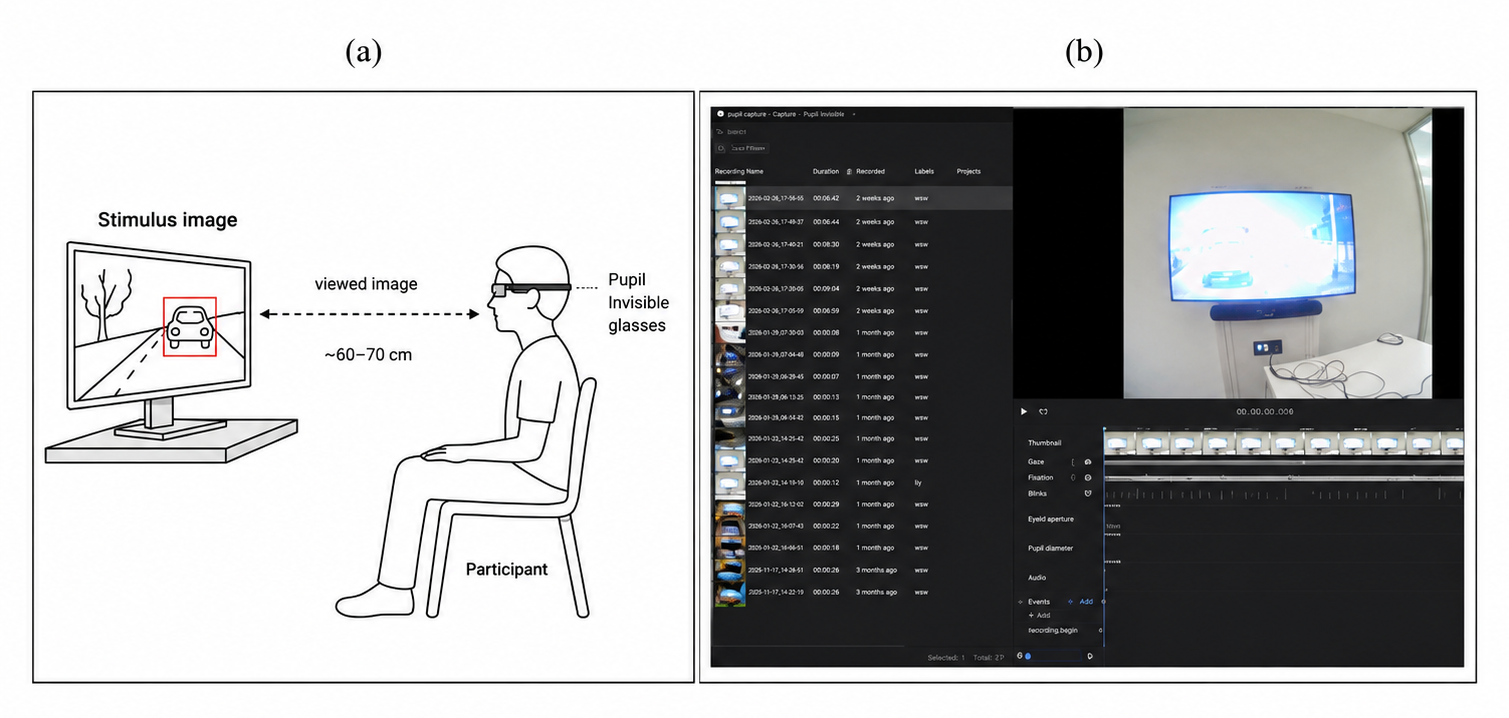}
\caption{Experimental setup. (a) Schematic representation of the eye-tracking protocol. Participants viewed stimulus images on a monitor while wearing Pupil Invisible glasses at a viewing distance of approximately 60--70 cm. (b) Example recording session in the Pupil Labs software environment showing the scene camera view and recorded eye-tracking signals.}
\label{fig:setup}
\end{figure}

\subsubsection{Apparatus}
The experimental setup is illustrated in Figure~\ref{fig:setup}(a). Participants were seated approximately 60--70 cm from a monitor displaying the stimulus images while wearing Pupil Invisible eye-tracking glasses. Eye movements were recorded continuously throughout the experiment using the Pupil Labs software environment.

Eye movements were recorded using Pupil Invisible glasses (Pupil Labs, Berlin, Germany), which capture binocular gaze at approximately 200 Hz alongside a forward-facing scene camera at 30 fps. The glasses do not require calibration and allow participants to view images naturally while providing reliable fixation, saccade, and pupil-size measurements \cite{PupilLabs2025,Kassner2014}.

\subsection{Gaze Data Processing}
Raw gaze data were recorded using the Pupil Invisible glasses and accessed through the Pupil Cloud platform, where recording sessions were reviewed and downloaded. Figure~\ref{fig:setup}(b) shows an example recording session, including the scene camera view and the eye-tracking signals captured by the Pupil Labs software environment. All files were stored on the Heriot-Watt University OneDrive. Gaze data were exported as CSV files containing timestamped gaze coordinates and worn-state flags, which served as input to the subsequent processing pipeline. 

\begin{figure}[H]
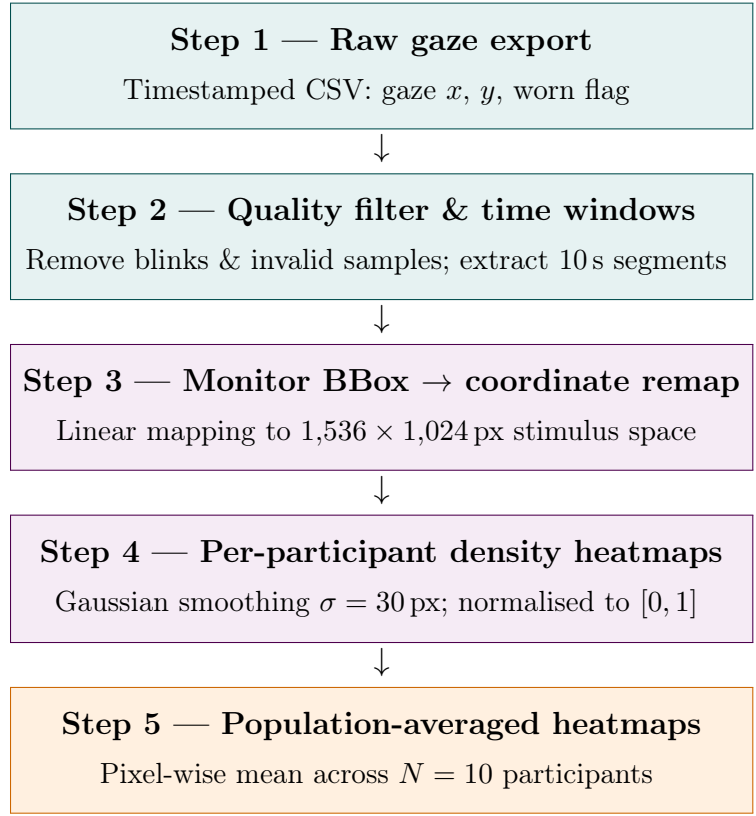

\centering
\begin{tabular}{c}
\fcolorbox{teal!60!black}{teal!10}{%
  \begin{minipage}{0.7\textwidth}
    \centering\medskip
    \textbf{Step 1 --- Raw gaze export}\\[4pt]
    {\small Timestamped CSV: gaze $x$, $y$, worn flag}
    \medskip
  \end{minipage}}\\[2pt]
$\downarrow$\\[2pt]
\fcolorbox{teal!60!black}{teal!10}{%
  \begin{minipage}{0.7\textwidth}
    \centering\medskip
    \textbf{Step 2 --- Quality filter \& time windows}\\[4pt]
    {\small Remove blinks \& invalid samples; extract 10\,s segments}
    \medskip
  \end{minipage}}\\[2pt]
$\downarrow$\\[2pt]
\fcolorbox{violet!60!black}{violet!8}{%
  \begin{minipage}{0.7\textwidth}
    \centering\medskip
    \textbf{Step 3 --- Monitor BBox $\rightarrow$ coordinate remap}\\[4pt]
    {\small Linear mapping to $1{,}536\times1{,}024$\,px stimulus space}
    \medskip
  \end{minipage}}\\[2pt]
$\downarrow$\\[2pt]
\fcolorbox{violet!60!black}{violet!8}{%
  \begin{minipage}{0.7\textwidth}
    \centering\medskip
    \textbf{Step 4 --- Per-participant density heatmaps}\\[4pt]
    {\small Gaussian smoothing $\sigma=30$\,px; normalised to $[0,1]$}
    \medskip
  \end{minipage}}\\[2pt]
$\downarrow$\\[2pt]
\fcolorbox{orange!80!black}{orange!12}{%
  \begin{minipage}{0.7\textwidth}
    \centering\medskip
    \textbf{Step 5 --- Population-averaged heatmaps}\\[4pt]
    {\small Pixel-wise mean across $N=10$ participants}
    \medskip
  \end{minipage}}\\
\end{tabular}
\caption{Human-side processing pipeline from exported gaze recordings to population-averaged heatmaps. Colour indicates processing stage: raw data acquisition (teal), spatial transformation (violet), and aggregated output (orange).}
\label{fig:human_pipeline}
\end{figure}

The processing pipeline consisted of five stages that are summarised in Figure~\ref{fig:human_pipeline}.

\paragraph{Filtering}
Samples were first filtered to retain only valid gaze data in Python using \texttt{pandas}. Rows with missing gaze coordinates were dropped. When the exported CSV contained a \texttt{worn} column, only samples with \texttt{worn}${}=1$ (device worn as intended) were retained. When a \texttt{blink id} column was present, samples with a non-missing blink identifier were excluded. These rules were applied programmatically rather than by manual inspection of individual rows. Timestamps in nanoseconds were converted to seconds and expressed relative to the earliest timestamp among the retained samples by subtracting that minimum value and dividing by $10^9$.

\paragraph{Temporal Segmentation}
The recording was segmented into 33 consecutive time windows, each corresponding to one stimulus image. Each window lasted 10 seconds, with a 2-second gap between windows. The first window began at $t = 6$\,s from the start of the recording to account for an initial setup period. Gaze samples within each window were extracted using a half-open interval to avoid
double-counting at window boundaries.

\paragraph{Bounding Box Annotation and Coordinate Mapping}
Because the Pupil Invisible scene camera records the participant's full field of view rather than the stimulus monitor alone, gaze coordinates in the exported CSV are expressed in scene-camera pixel space and must be mapped onto the stimulus image. For each 10-second window, the temporal midpoint was computed from the same onset--offset schedule used for gaze segmentation. The corresponding frame in the exported scene-camera video was then retrieved with a custom Python script using OpenCV: the script seeks to the frame index associated with the midpoint timestamp given the video frame rate. On each retrieved frame, the visible monitor region was outlined manually by clicking the top-left and bottom-right corners of the screen; these two points define an axis-aligned bounding box. The corner coordinates were saved to a CSV file for reuse in subsequent steps. During heatmap generation, gaze samples were retained only if they fell inside the bounding box and were linearly remapped to the stimulus resolution of $1{,}536 \times 1{,}024$ pixels.

\paragraph{Heatmap Generation}
For each time window, the remapped gaze coordinates were used to construct a fixation density map. Each gaze point was rendered as a filled circle with a radius of 7 pixels on a blank canvas. Gaussian smoothing with $\sigma = 30$ pixels was applied to produce a continuous spatial distribution, and the resulting map was normalised to the range $[0, 1]$. This process was applied independently for each participant and each window, producing 330 individual gaze heatmaps in total (10 participants $\times$ 33 images).

\begin{figure}[H]
\centering
\begin{tabular}{cc}
  \includegraphics[width=0.38\textwidth]{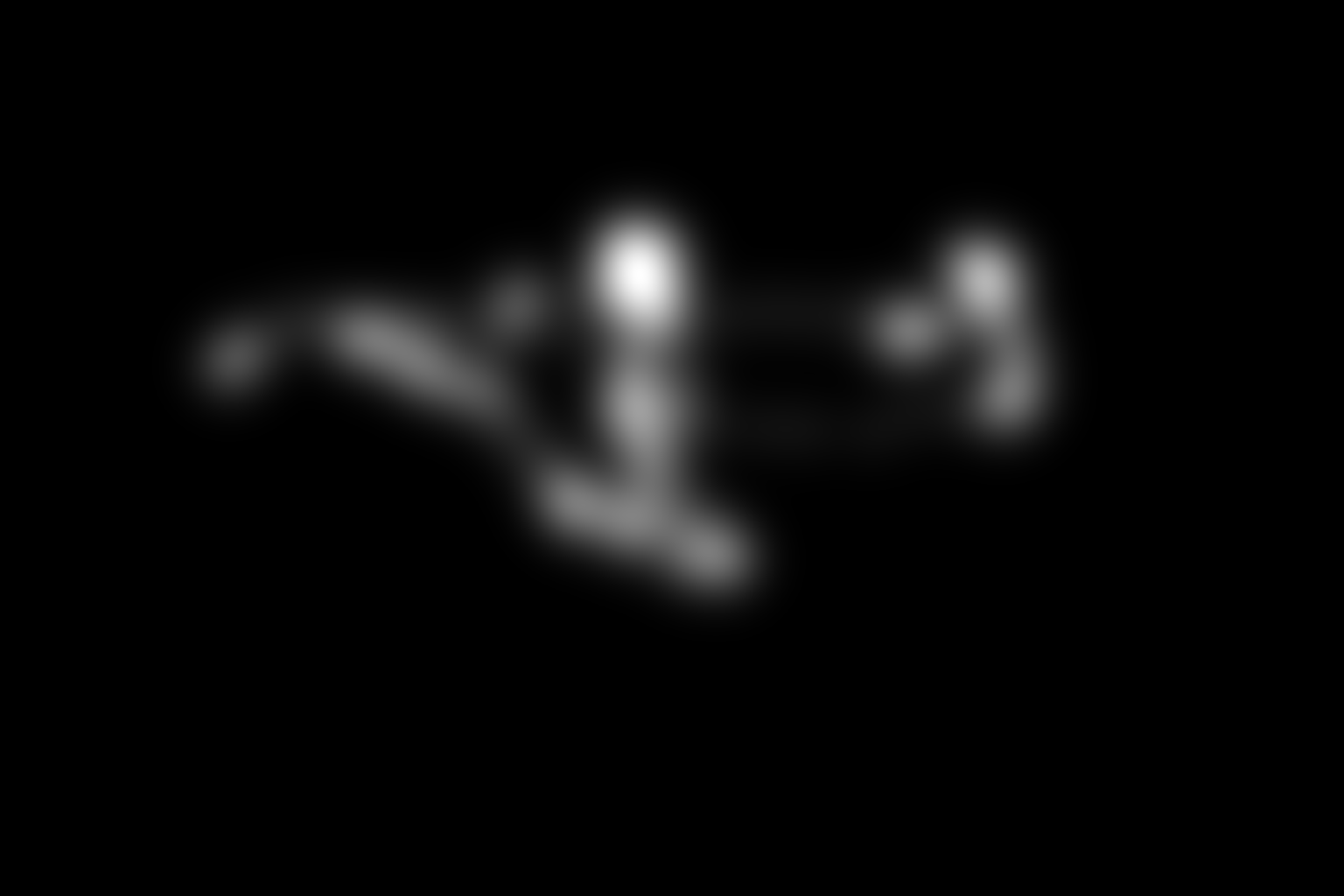} &
  \includegraphics[width=0.38\textwidth]{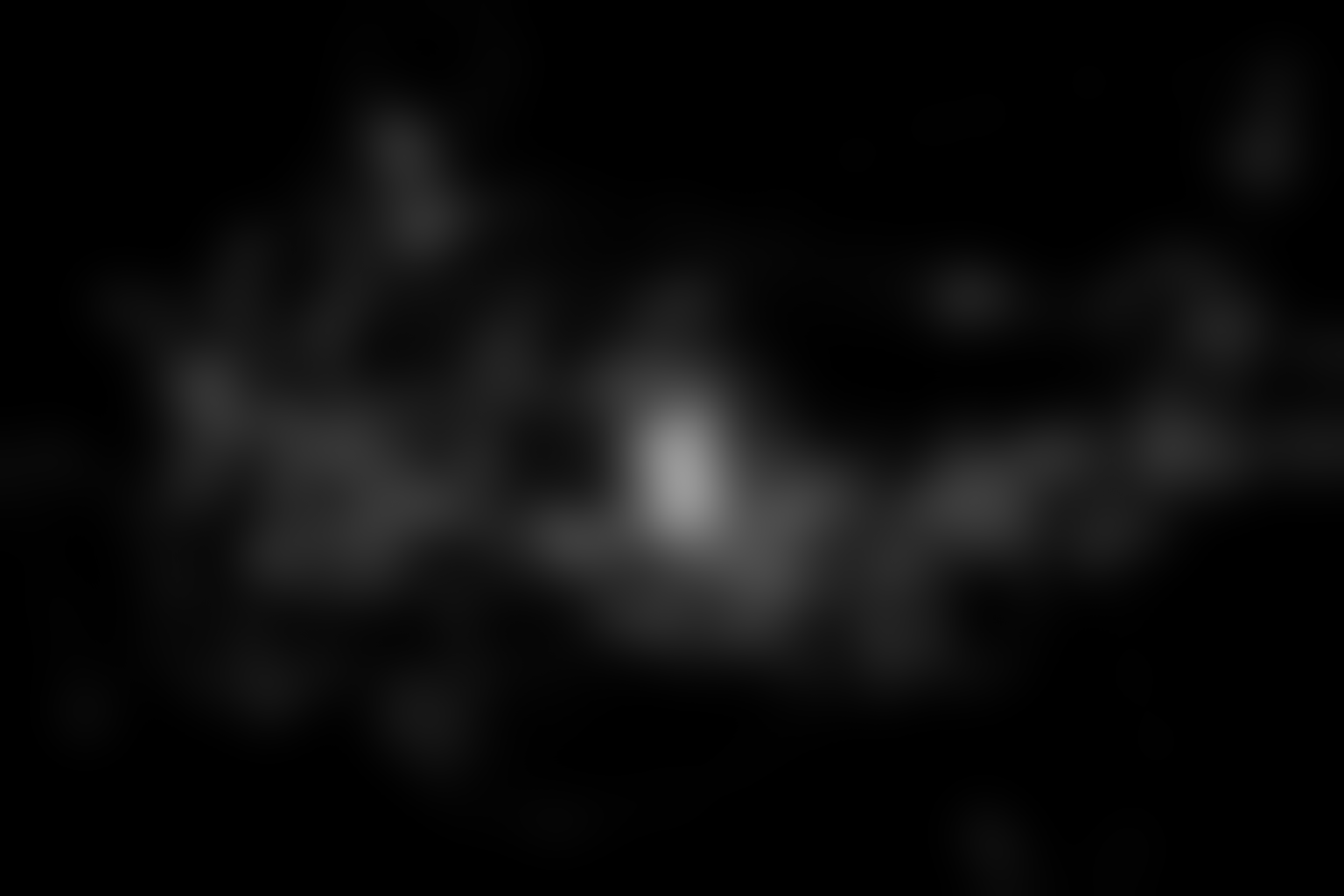} \\
  (a) Example individual gaze heatmap 1 &
  (b) Example individual gaze heatmap 2 \\[4pt]
  \includegraphics[width=0.38\textwidth]{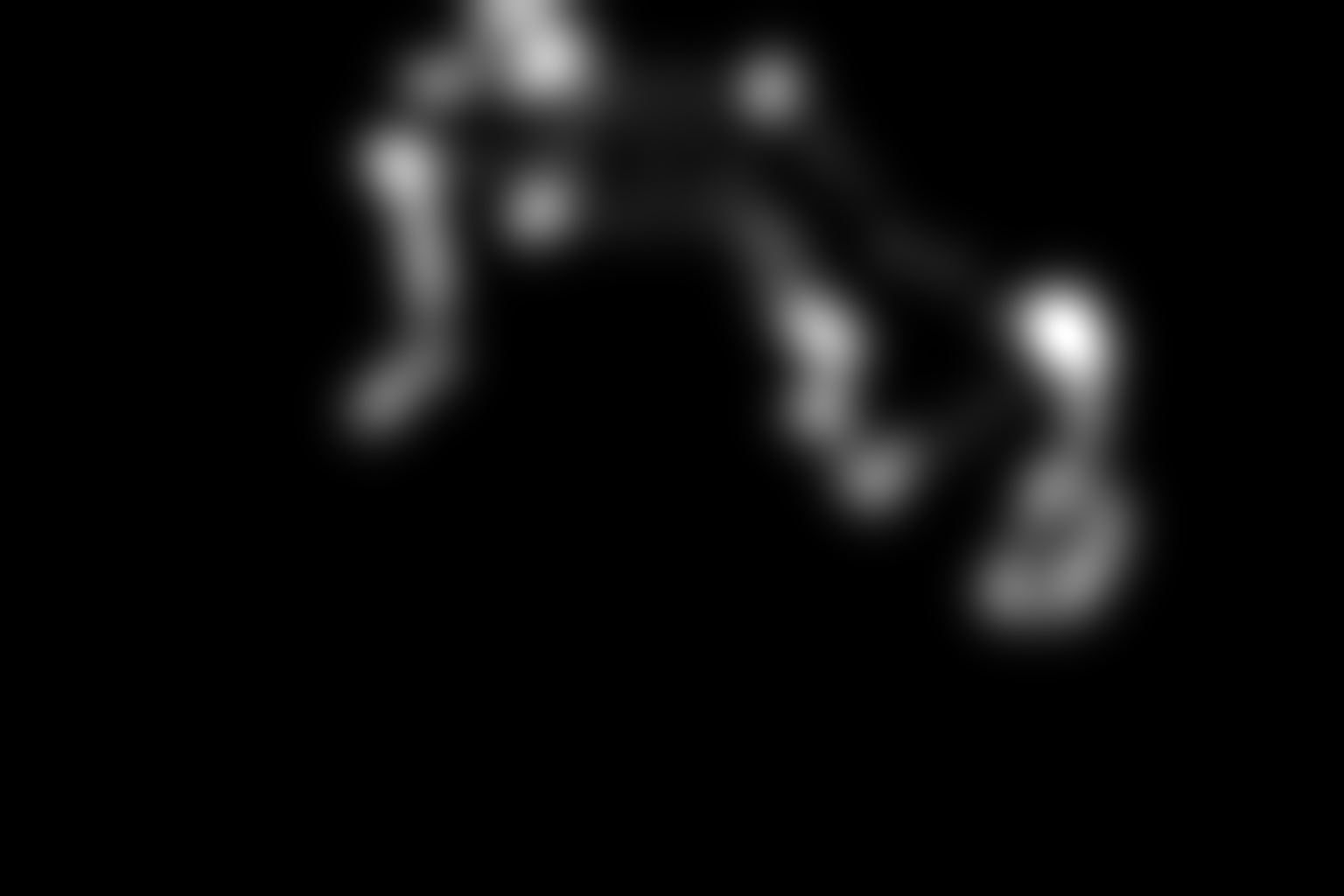} &
  \includegraphics[width=0.38\textwidth]{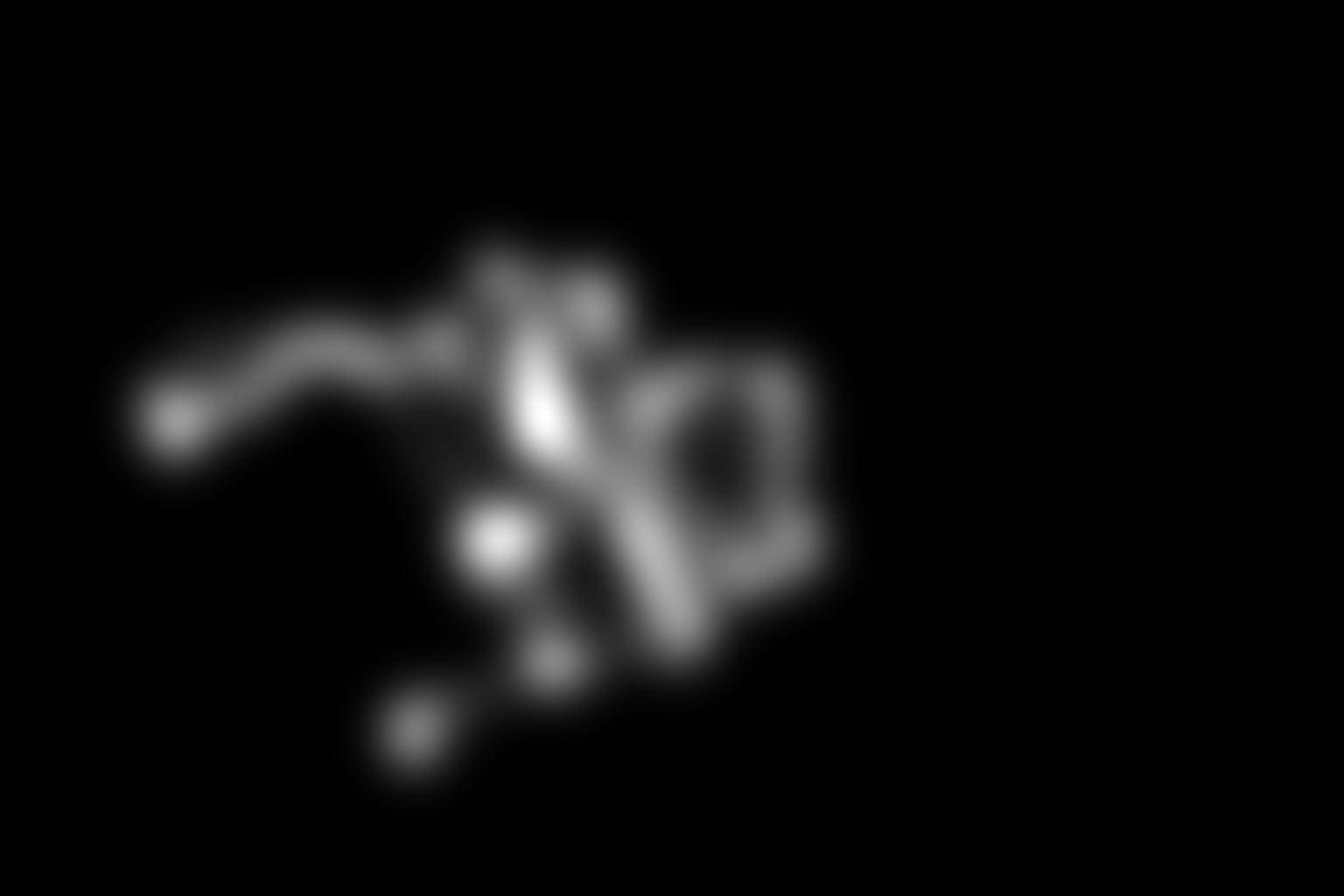} \\
  (c) Example individual gaze heatmap 3 &
  (d) Example individual gaze heatmap 4 \\[4pt]
  \includegraphics[width=0.38\textwidth]{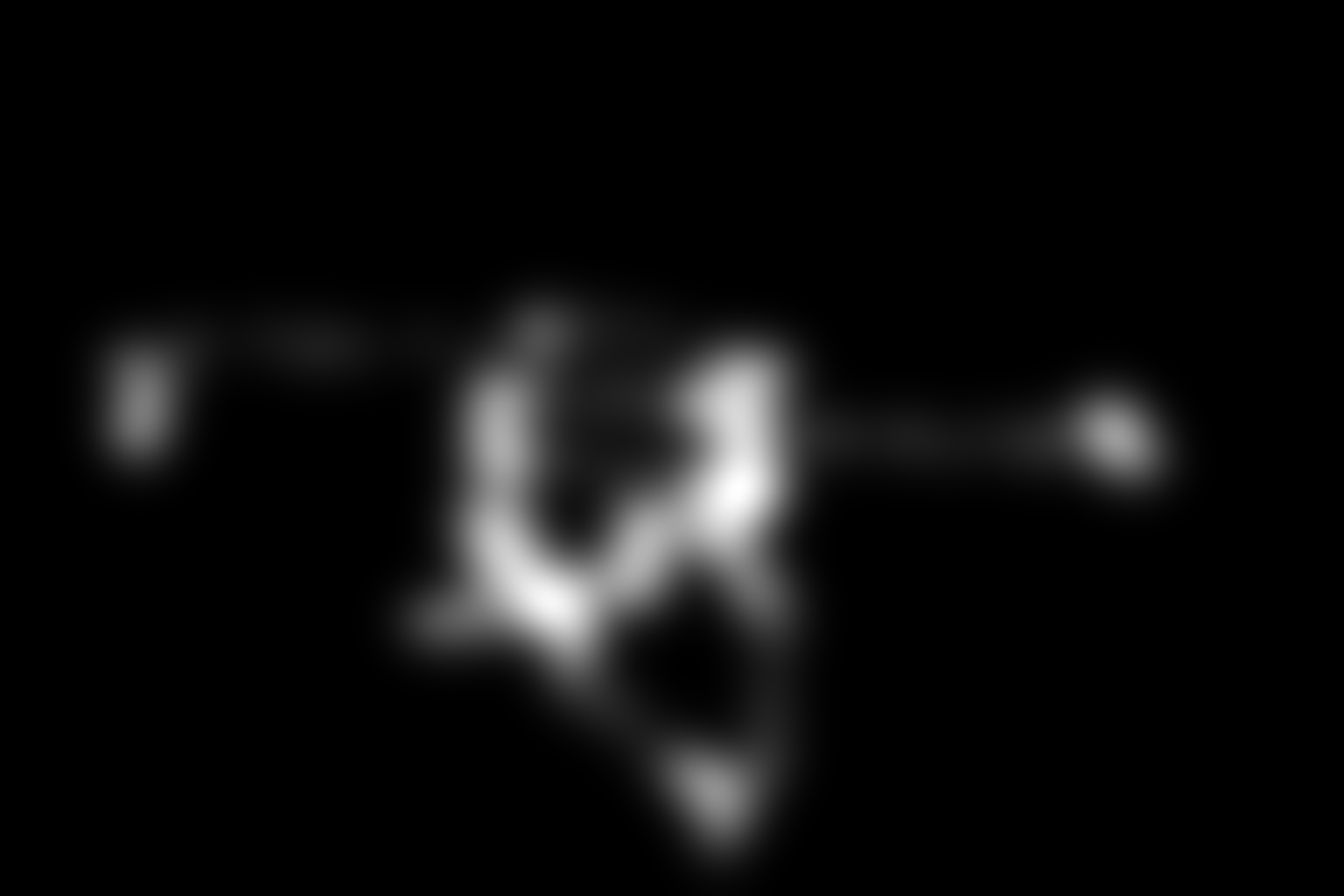} &
  \includegraphics[width=0.38\textwidth]{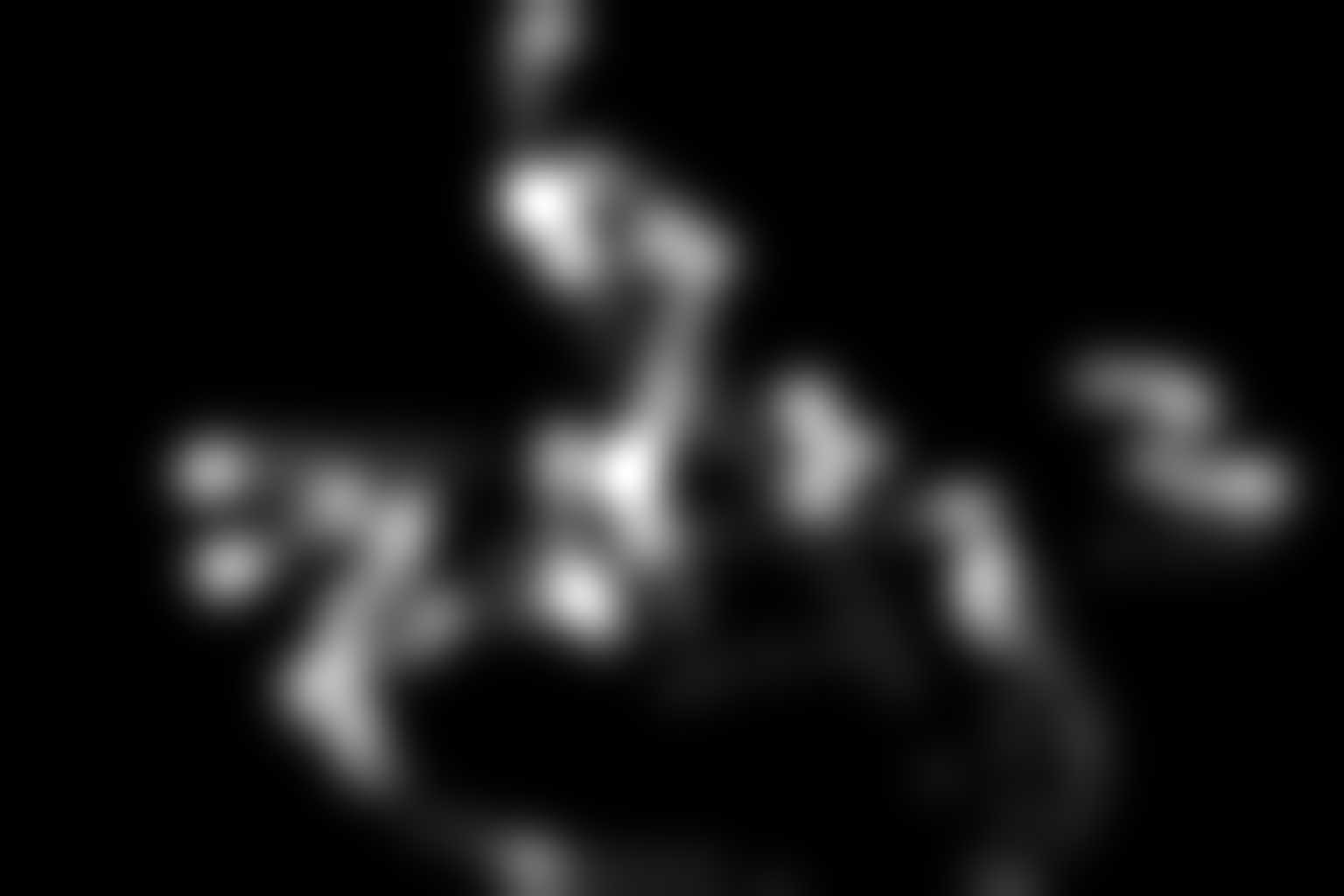} \\
  (e) Example individual gaze heatmap 5 &
  (f) Example individual gaze heatmap 6 \\[4pt]
  \includegraphics[width=0.38\textwidth]{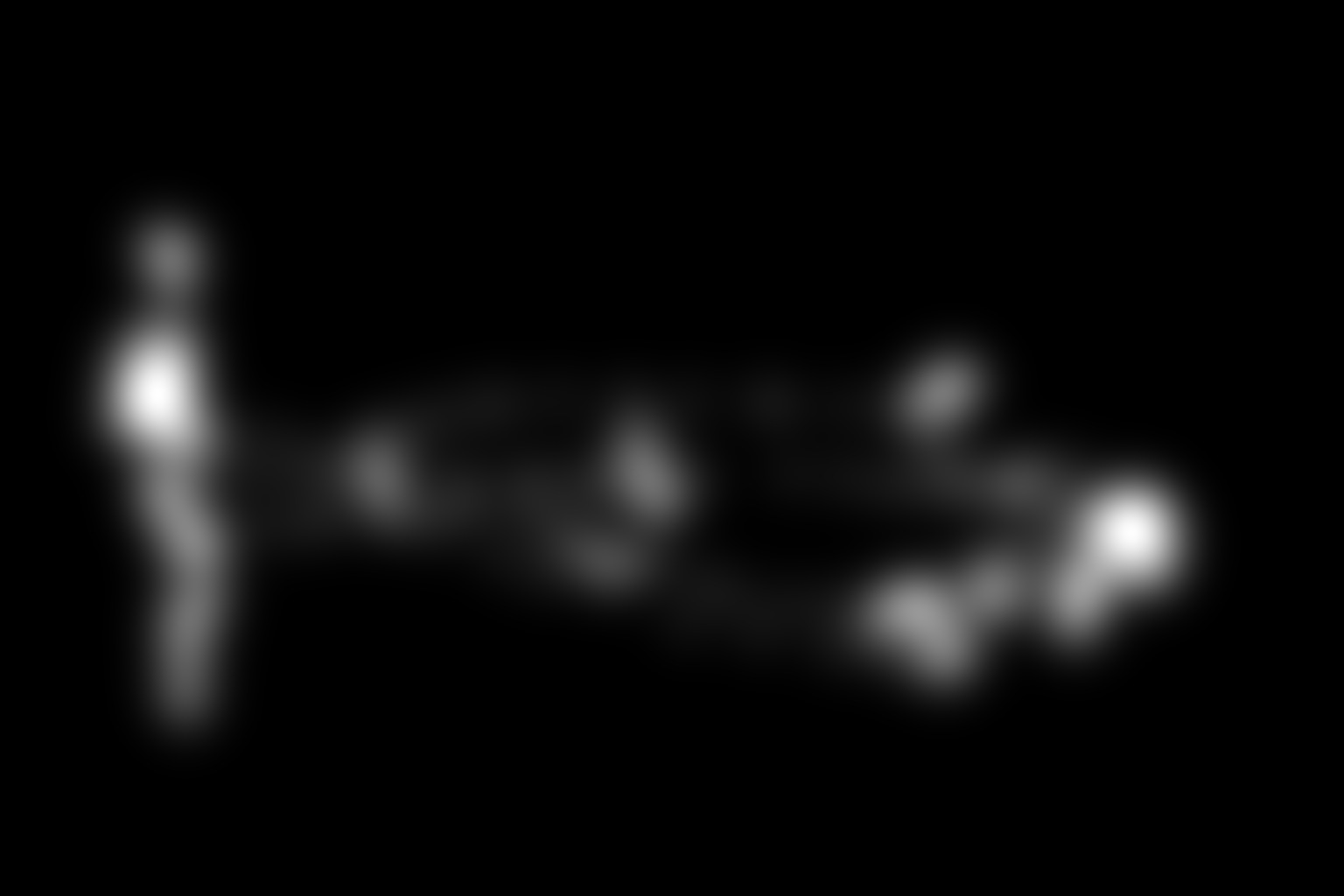} &
  \includegraphics[width=0.38\textwidth]{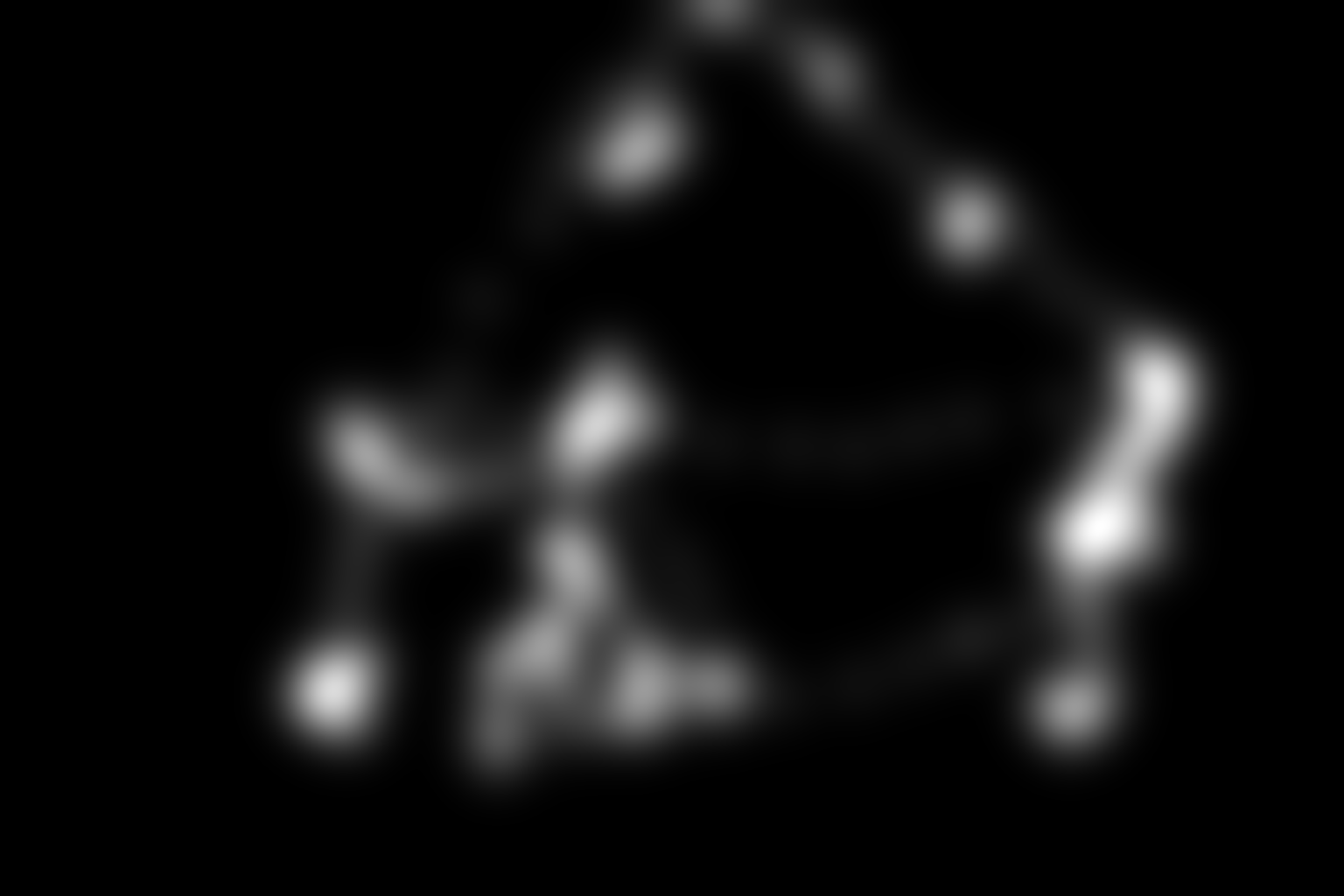} \\
  (g) Example individual gaze heatmap 7 &
  (h) Example individual gaze heatmap 8 \\
\end{tabular}
\caption{Examples of individual human gaze heatmaps generated from fixation data. \label{fig:human_heatmaps}}
\end{figure}

The 10 individual heatmaps for each window were then averaged pixel-wise to produce 33 population-level gaze heatmaps, matched by window index to ensure consistent alignment across participants regardless of differences in recording start times.

\begin{figure}[H]
\centering
\begin{tabular}{cc}
  \includegraphics[width=0.45\textwidth]{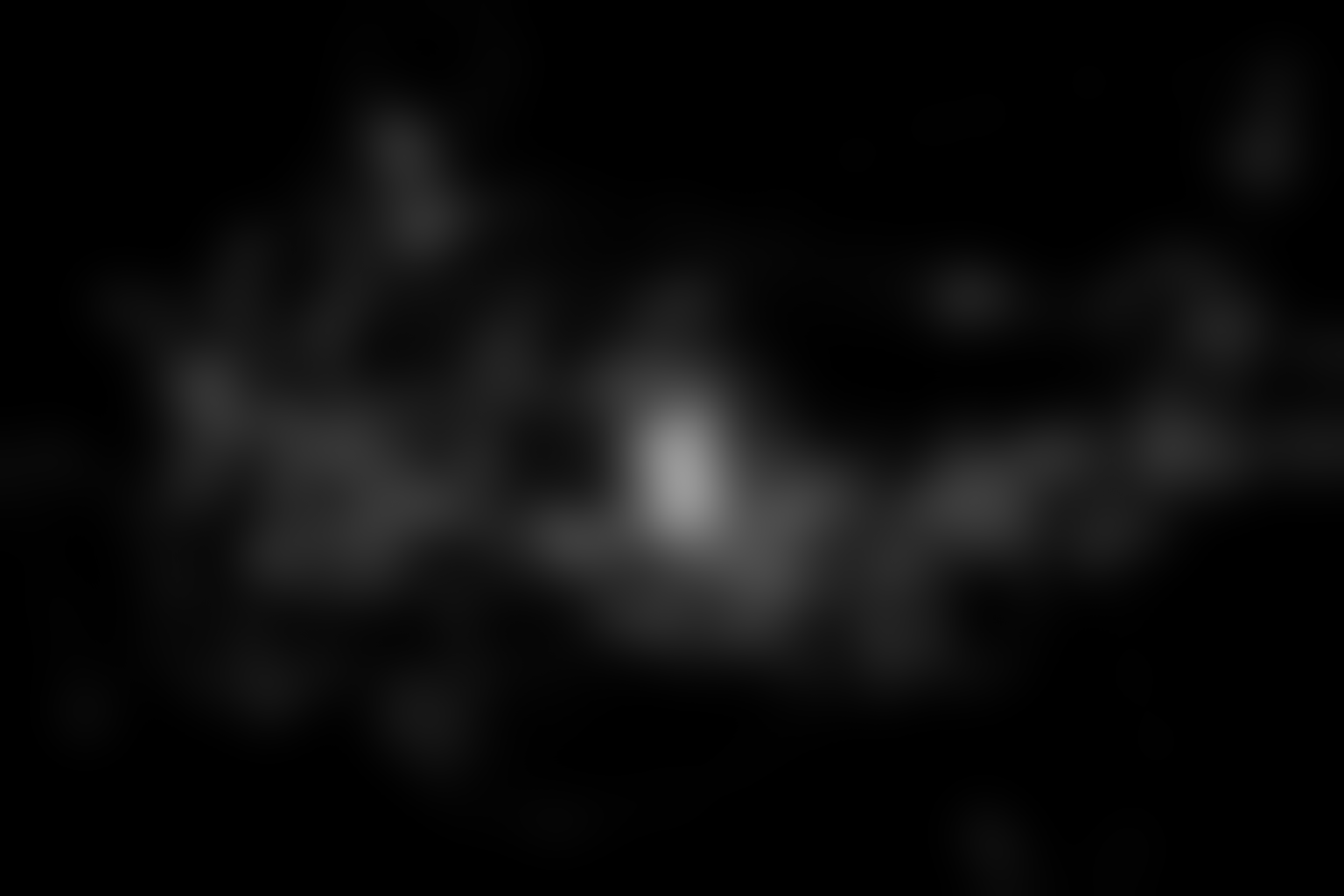} &
  \includegraphics[width=0.45\textwidth]{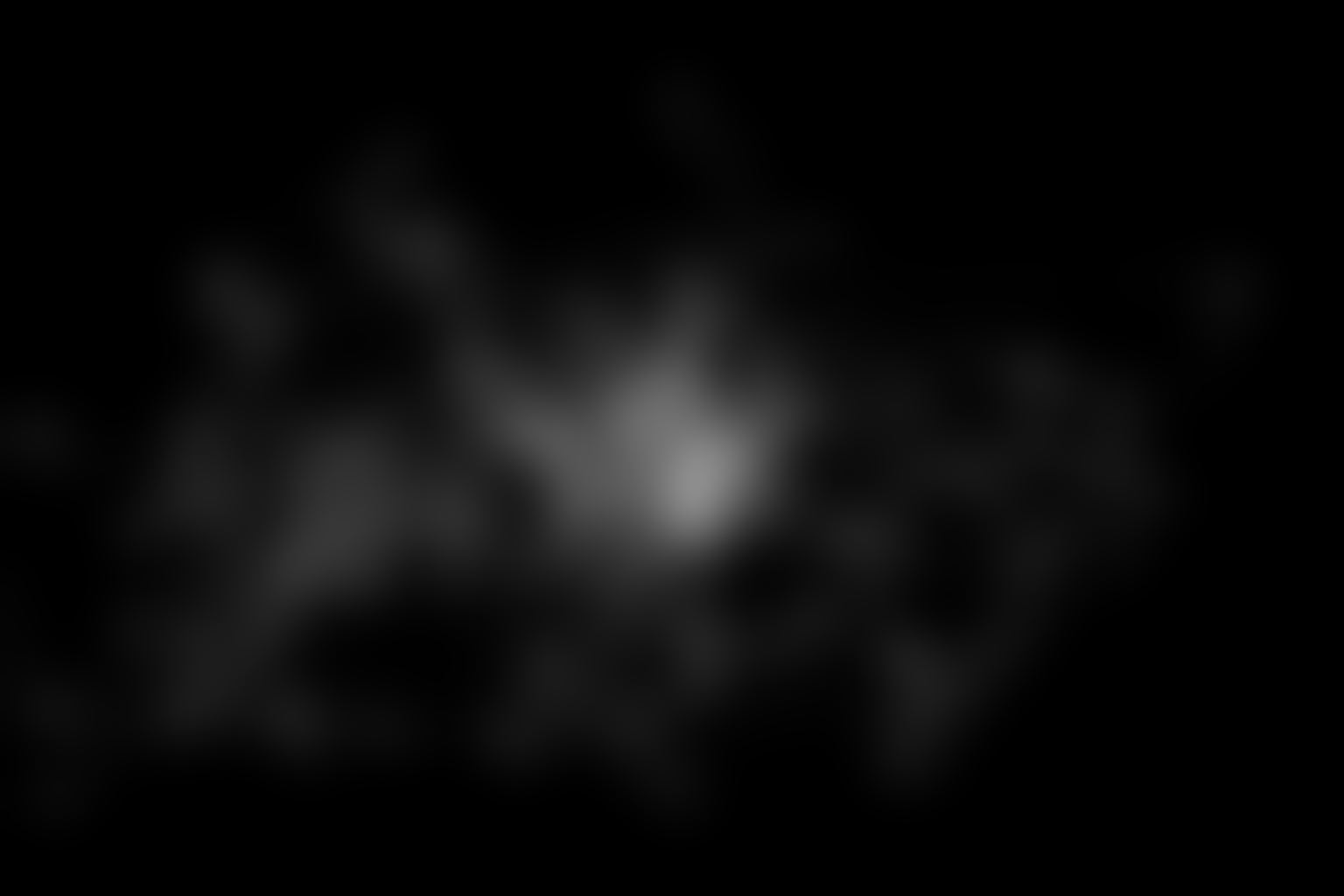} \\
  (a) Population-averaged gaze heatmap 1 &
  (b) Population-averaged gaze heatmap 2 \\[2pt]
  \includegraphics[width=0.45\textwidth]{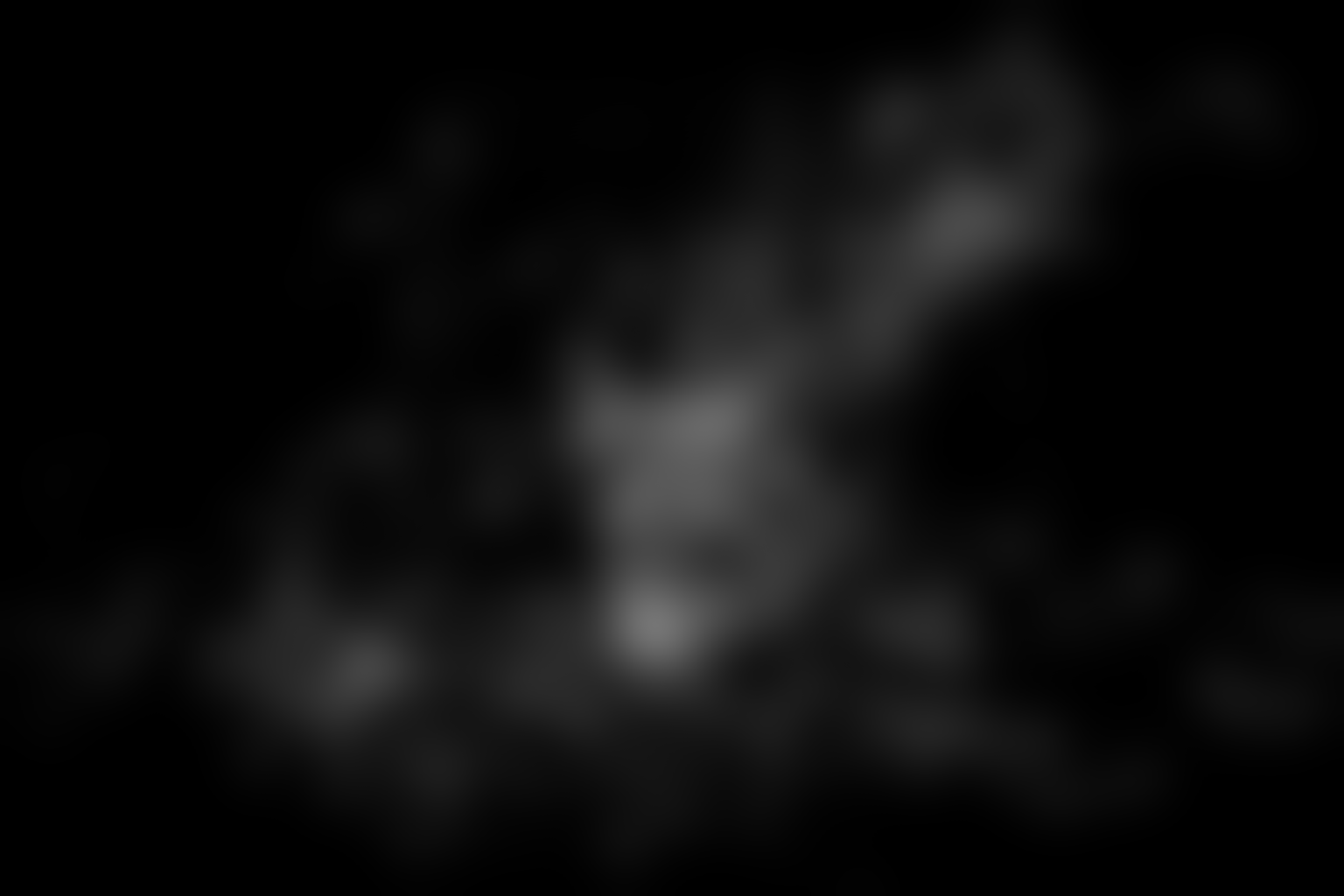} &
  \includegraphics[width=0.45\textwidth]{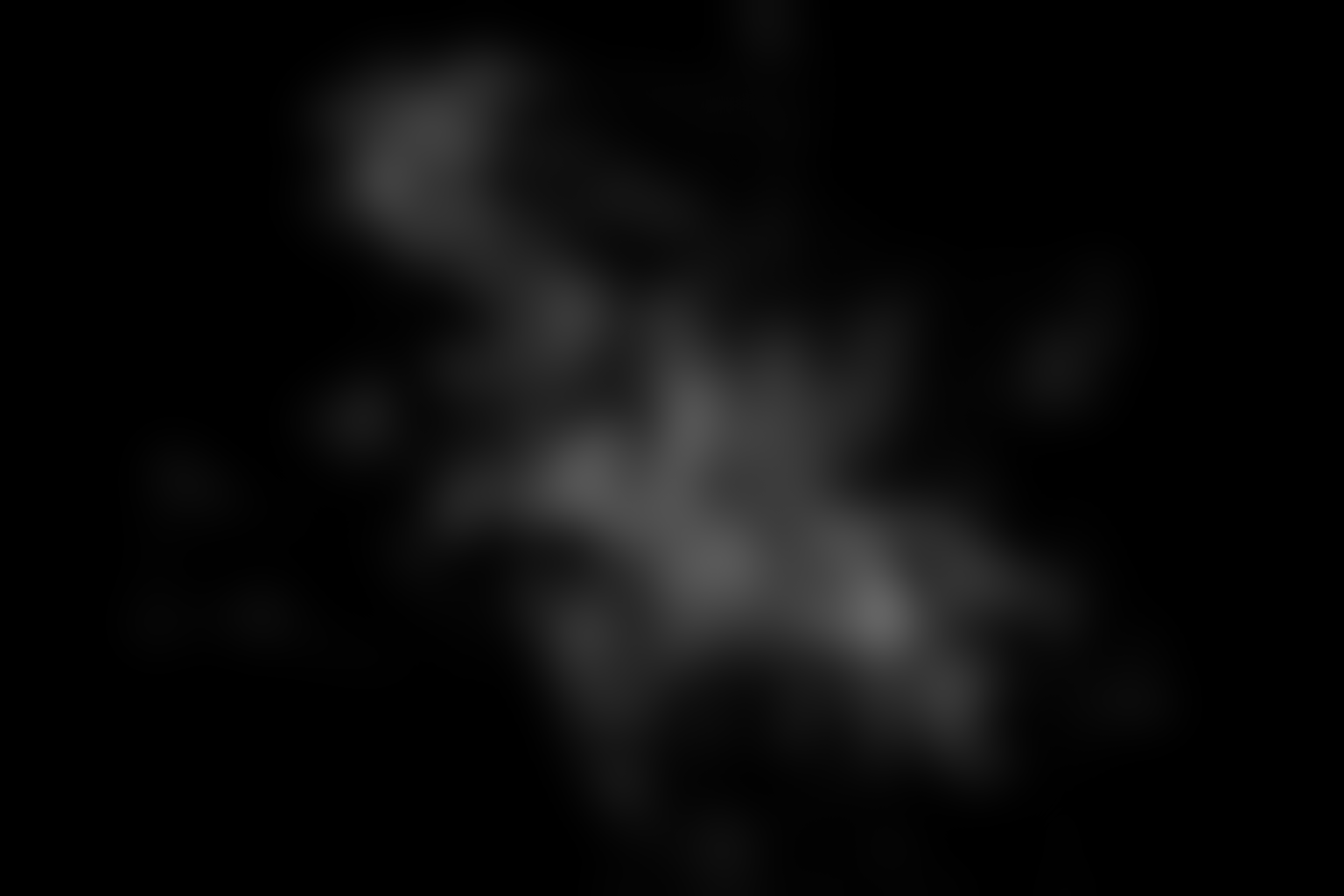} \\
  (c) Population-averaged gaze heatmap 3 &
  (d) Population-averaged gaze heatmap 4 \\
\end{tabular}
\caption{Population-averaged gaze heatmaps obtained by pixel-wise averaging across all participants.
\label{fig:population_heatmap}}
\end{figure}

\subsection{Dataset Summary}
The dataset spans four everyday environment types with varied levels of potential risk. With ten participants each viewing all 33 stimuli, the dataset yields 330 individual gaze heatmaps and 33 population-averaged heatmaps, which serve as the human ground truth for the AI comparison. Table~\ref{tab:dataset_summary} summarises the key properties of the stimuli, participant sample, and recorded gaze data.

\begin{table}[H]
\centering
\caption{Summary of the collected eye-tracking dataset.}
\label{tab:dataset_summary}
\begin{tabularx}{\textwidth}{lX}
\toprule
\textbf{Property} & \textbf{Detail} \\
\midrule
Total stimuli & 33 scene images \\
Scene coverage & A diverse set of everyday environments \\
Image resolution & $1{,}536 \times 1{,}024$ pixels \\
Display duration & 10\,s per image; 2\,s inter-stimulus interval \\
Presentation screen & 24-inch monitor at 60--70\,cm viewing distance \\
Total participants & 10 university students (6 female, 4 male) \\
Vision & Normal or corrected-to-normal \\
Ethics approval & MACS Ethics Committee, Heriot-Watt University \\
Total recording sessions & 330 ($10 \times 33$) \\
Sampling rate & $\sim$200\,Hz (binocular) \\
Scene camera frame rate & 30\,fps \\
Total session duration & $\sim$10\,min per participant \\
Gaze data format & Timestamped CSV (gaze coordinates, fixation ID, blink ID) \\
Individual heatmaps & 330 (one per participant per stimulus) \\
Population-averaged heatmaps & 33 (one per stimulus, averaged across all participants) \\
\bottomrule
\end{tabularx}
\end{table}

\subsection{Artificial Intelligence Saliency Map Generation}
\label{sec:ai_pipeline}

\subsubsection{Model Selection}
GPT-4o was selected as the vision-language model because it supports high-resolution image inputs and demonstrates strong performance on multimodal reasoning tasks, including scene understanding, visual question answering, and structured interpretation of visual content \cite{OpenAI2024HelloGPT4o}. These capabilities make it well suited to generating spatial descriptions of scene content that can be transformed into saliency maps and compared with human gaze behaviour. Unlike classical saliency networks trained directly on fixation datasets, GPT-4o can be prompted to produce a structured JSON representation of spatial emphasis; the Python implementation converts this output into a continuous saliency map for comparison with aggregated human gaze data.

\subsubsection{Prompt Design}
\label{sec:prompt_design}
A single prompt for AI inference was implemented in Python and sent with each image to GPT-4o. The prompt does \emph{not} use a fixed spatial grid: it asks the model to return \emph{only} a JSON object (no markdown fences) containing a \texttt{fixations} array. Each element has normalised coordinates $x,y \in [0,1]$ and a saliency weight $w \in [0.1,1.0]$. The model is instructed to act as a visual-saliency expert, to predict where human gaze would fall during ten seconds of free viewing, and to provide roughly twenty to twenty-five points, spread across salient objects and regions (including hazards, faces, vehicles, text), with a mild centre bias. The verbatim prompt is provided in Appendix~\ref{app:prompt}.

\subsubsection{Inference Pipeline and Post-Processing}

\begin{figure}[H]
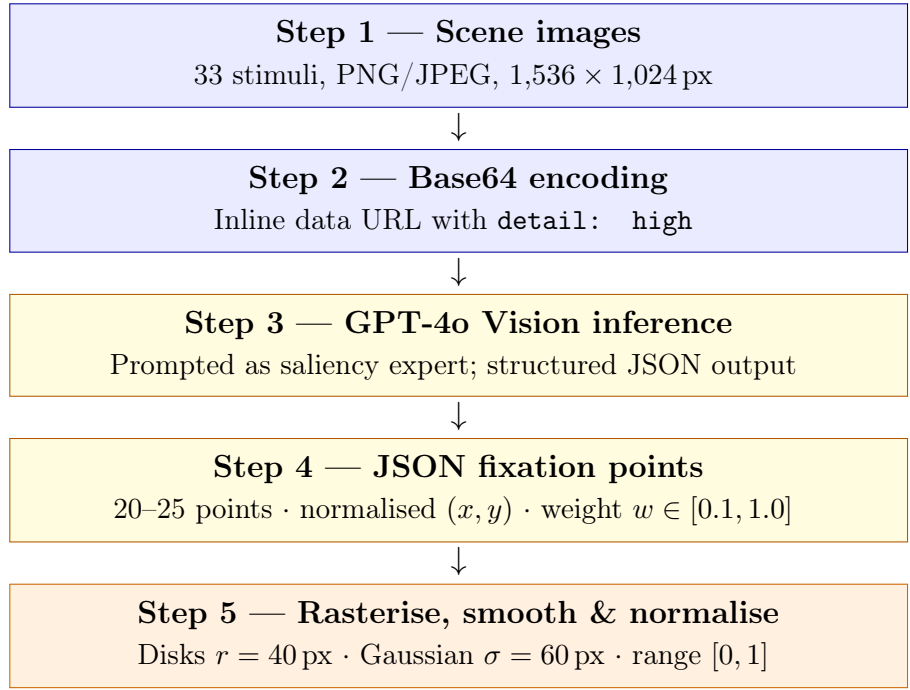

\centering
\begin{tabular}{c}
\fcolorbox{blue!60!black}{blue!8}{%
  \begin{minipage}{0.85\textwidth}
    \centering\smallskip
    \textbf{Step 1 --- Scene images}\\[1pt]
    {\small 33 stimuli, PNG/JPEG, $1{,}536\times1{,}024$\,px}
    \smallskip
  \end{minipage}}\\[1pt]
$\downarrow$\\[1pt]
\fcolorbox{blue!60!black}{blue!8}{%
  \begin{minipage}{0.85\textwidth}
    \centering\smallskip
    \textbf{Step 2 --- Base64 encoding}\\[1pt]
    {\small Inline data URL with \texttt{detail: high}}
    \smallskip
  \end{minipage}}\\[1pt]
$\downarrow$\\[1pt]
\fcolorbox{orange!70!black}{yellow!15}{%
  \begin{minipage}{0.85\textwidth}
    \centering\smallskip
    \textbf{Step 3 --- GPT-4o Vision inference}\\[1pt]
    {\small Prompted as saliency expert; structured JSON output}
    \smallskip
  \end{minipage}}\\[1pt]
$\downarrow$\\[1pt]
\fcolorbox{orange!70!black}{yellow!15}{%
  \begin{minipage}{0.85\textwidth}
    \centering\smallskip
    \textbf{Step 4 --- JSON fixation points}\\[1pt]
    {\small 20--25 points $\cdot$ normalised $(x,y)$ $\cdot$ weight $w\in[0.1,1.0]$}
    \smallskip
  \end{minipage}}\\[1pt]
$\downarrow$\\[1pt]
\fcolorbox{orange!80!black}{orange!12}{%
  \begin{minipage}{0.85\textwidth}
    \centering\smallskip
    \textbf{Step 5 --- Rasterise, smooth \& normalise}\\[1pt]
    {\small Disks $r=40$\,px $\cdot$ Gaussian $\sigma=60$\,px $\cdot$ range $[0,1]$}
    \smallskip
  \end{minipage}}\\
\end{tabular}
\caption{AI saliency map generation pipeline: GPT-4o returns structured JSON fixation points; the pipeline rasterises weighted disks, applies Gaussian smoothing ($\sigma=60$\, px), and normalises to produce maps comparable with human gaze heatmaps. Colour indicates pipeline stage: image preparation (blue), model inference (amber), and spatial rendering (orange).}
\label{fig:ai_pipeline}
\end{figure}

The AI saliency map generation pipeline is shown in Figure~\ref{fig:ai_pipeline}.

The scientific aim of this AI pipeline is to convert each stimulus image into a full-resolution continuous saliency map that can be compared pixel-wise with the population gaze heatmaps, without training a separate neural network. The implementation aim is to use the official Python \texttt{openai} package: it provides the \texttt{OpenAI} client and \texttt{chat.completions.create} so that each call programmatically builds the multimodal \texttt{messages} payload (image plus text), sends it to the hosted GPT-4o endpoint, and returns the model's text for downstream parsing, avoiding ad hoc HTTP code while staying aligned with the current API specification \cite{OpenAI2024HelloGPT4o}.

A fully automated script was written around the \texttt{openai} library to process every image in the stimulus folder (33 images in this study) through GPT-4o and to turn responses into normalised saliency heatmaps. Authentication uses the \texttt{OPENAI\_API\_KEY} environment variable; the Python client attaches the same credential that a REST client would send in the \texttt{Authorization: Bearer} header.

Each stimulus image was Base64-encoded and sent through the OpenAI Vision API together with the structured prompt from Section~\ref{sec:prompt_design} (verbatim text in Appendix~\ref{app:prompt}). Following the API format, the \texttt{user} message \texttt{content} lists an \texttt{image\_url} entry (\texttt{detail: high}) and then a \texttt{text} entry. Requests used \texttt{client.chat.completions.create} with \texttt{model="gpt-4o"} and were issued sequentially with short delays to respect rate limits. To save cost during development, the script skips the API call when a raw JSON response already exists for an image, and skips rasterisation when a normalised grayscale heatmap already exists. Raw model outputs were stored under \texttt{openai\_raw\_v4/} so heatmaps could be regenerated after rendering changes without new API calls.

Response text was stripped of any markdown code fences and parsed as JSON. Each $(x,y)$ was mapped from normalised coordinates to pixel indices on a $1{,}536 \times 1{,}024$ canvas. For each point, a filled disk of radius $r = 40$ pixels was drawn with intensity $w$ using OpenCV
(\texttt{cv2.circle}), accumulating a sparse density map. This implementation does \emph{not} use a low-resolution grid. Therefore, bicubic interpolation is not applied. Spatial continuity is obtained from weighted disk rasterisation followed by Gaussian smoothing.

\subsubsection{Smoothing and Normalisation}
\label{sec:api_smooth_norm}

Gaussian smoothing with $\sigma = 60$\, px was applied to the rasterised map using \texttt{scipy.ndimage.gaussian\_filter}, producing a smooth saliency field. An isotropic Gaussian kernel in continuous form satisfies:

\begin{equation}
G_{\sigma}(x,y) =
\frac{1}{2\pi\sigma^{2}}
\exp\!\left(-\frac{x^{2}+y^{2}}{2\sigma^{2}}\right),
\label{eq:gaussian}
\end{equation}

where $G_{\sigma}(x,y)$ denotes the Gaussian kernel value at spatial coordinates $(x,y)$ relative to the kernel centre, and $\sigma$ is the standard deviation controlling the spread of the Gaussian distribution. Larger values of $\sigma$ produce greater smoothing by distributing the influence of each fixation over a wider spatial region.

This is equivalent to applying a separable one-dimensional Gaussian with standard deviation $\sigma$ along the rows and columns of the discrete image grid.

An even larger $\sigma$ was used here than in the human heatmap pipeline ($\sigma = 30$ pixels) because the AI branch starts from 20--25 sparse weighted fixation points rather than the dense continuous gaze samples available in the human pipeline; a larger kernel better approximates the spatial spread expected when aggregating population-level attention patterns.

\subsection{Evaluation Metrics}
\label{sec:metrics}
This section presents the evaluation of the spatial alignment between the GPT-4o AI risk heatmaps and the population-averaged human gaze heatmaps across all 33 stimulus images. It begins by defining the four saliency metrics used for comparison, followed by quantitative results, qualitative visualisation, and discussion of the findings. Four complementary saliency
metrics were used \cite{Bylinskii2019SaliencyMetrics,Kummerer2018SaliencyBenchmarking}.

\subsubsection{Pearson Correlation}
Pearson Correlation $r$ measures the linear pixel-wise correlation between the AI heatmap and the population-averaged human gaze heatmap after mean-centring both maps, and is defined as:

\begin{equation}
r = \frac{\displaystyle\sum_{i}(A_i - \bar{A})(B_i - \bar{B})}
{\sqrt{\displaystyle\sum_{i}(A_i - \bar{A})^2}
 \cdot \sqrt{\displaystyle\sum_{i}(B_i - \bar{B})^2}}
\end{equation}

where $A_i$ and $B_i$ denote the pixel values of the AI heatmap and the human gaze heatmap respectively, and $\bar{A}$, $\bar{B}$ are their spatial means. The result is a value in the range $[-1, 1]$, where 1 indicates perfect positive correlation, 0 indicates no linear relationship, and negative values indicate inverse correspondence. Pearson $r$ captures the degree to which the global spatial structure of the two maps is similar, and provides an intuitive measure of overall spatial agreement.

\subsubsection{Normalised Scanpath Saliency}
Normalised Scanpath Saliency (NSS) evaluates the AI heatmap specifically at pixel locations where human gaze was recorded. The AI map is first normalised to zero mean and unit variance, and NSS is formally defined as:

\begin{equation}
\text{NSS} = \frac{1}{N} \sum_{i \in \mathcal{F}} \hat{s}_i,
\qquad \hat{s}_i = \frac{s_i - \mu_s}{\sigma_s}
\end{equation}
where $\hat{s}_i$ is the z-scored value of the AI saliency map at pixel $i$, $\mathcal{F}$ denotes the set of fixated pixels (defined as pixels where the human gaze density exceeds a threshold of 0.1), $N = |\mathcal{F}|$ is the total number of fixated pixels, and $\mu_s$,
$\sigma_s$ are the mean and standard deviation of the AI map \cite{Bylinskii2019SaliencyMetrics}.

A positive NSS score indicates that the AI map assigns above-average values at fixated locations. A score near zero indicates no correspondence, and a negative score indicates that the AI map assigns below-average values where humans looked. NSS has no fixed theoretical upper bound; values above 1.0 indicate meaningful alignment, and human inter-observer consistency typically yields NSS values in the range of 1--3 \cite{Kummerer2018SaliencyBenchmarking}.

\subsubsection{Kullback--Leibler Divergence}
Kullback--Leibler (KL) divergence measures the distributional difference between the two heatmaps by treating each as a probability distribution over image pixels, and is defined as:

\begin{equation}
D_{\mathrm{KL}}(H \| A) = \sum_{i} H_i \log \frac{H_i}{A_i}
\end{equation}

where $H_i$ and $A_i$ are the normalised pixel values of the human gaze map and the AI heatmap, respectively \cite{Bylinskii2019SaliencyMetrics}. A small epsilon of $10^{-10}$ is added to all pixel values before normalisation to prevent division by zero when computing the log ratio. Both maps are normalised so that their values sum to one, and $D_{\mathrm{KL}}$ is computed using \texttt{scipy.stats.entropy}.

KL divergence has a lower bound of zero, which indicates identical distributions; there is no fixed upper bound, and larger values indicate greater spatial disagreement. In this study, lower KL divergence indicates that the AI heatmap more closely approximates the spatial distribution of human visual attention. KL divergence is asymmetric and is particularly sensitive to cases where the AI map assigns very low probability to regions that receive high human attention \cite{Bylinskii2019SaliencyMetrics}.

\subsubsection{Area Under the Receiver Operating Characteristic Curve (AUC-Judd)}
AUC-Judd treats the comparison as a binary classification problem. The human gaze map is thresholded at 0.1 to produce a binary fixation map. The AI heatmap is used as a continuous ranking score, and a Receiver Operating Characteristic curve is constructed by sweeping 100 evenly spaced thresholds across the AI map's value range and computing the true positive rate and false positive rate at each threshold \cite{Kummerer2018SaliencyBenchmarking}. Boundary points at $(0,0)$ and $(1,1)$ are appended, and the area under the resulting curve is computed using the trapezoidal rule via \texttt{numpy.trapezoid}. A value of 1.0 indicates perfect discrimination of fixated from non-fixated locations, while 0.5 corresponds to chance
performance.

\section{Results}
\label{sec:results}

\subsection{Quantitative Results}
\begin{figure}[H]
\centering
\begin{tabular}{cc}
  \includegraphics[width=0.45\textwidth]{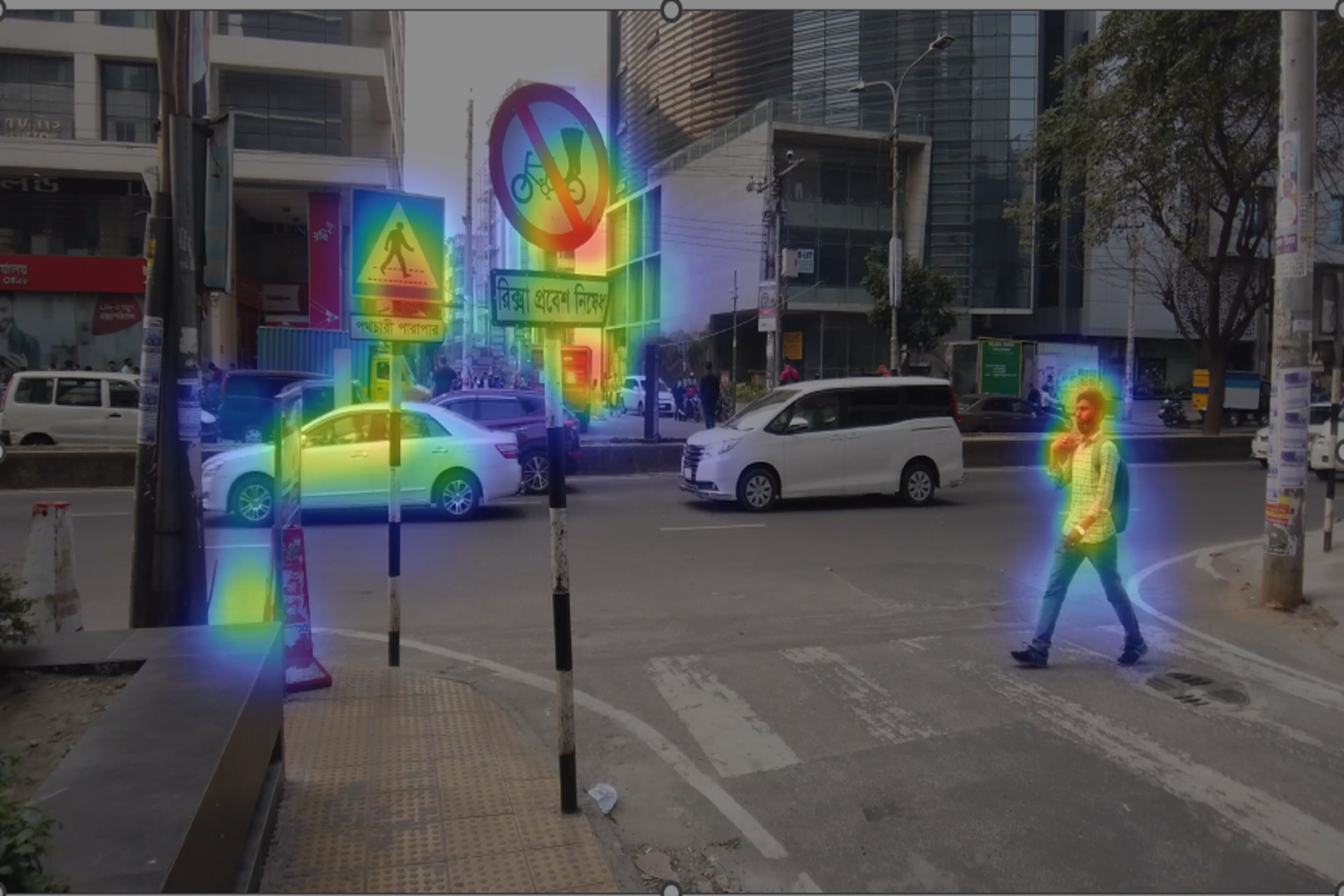} &
  \includegraphics[width=0.45\textwidth]{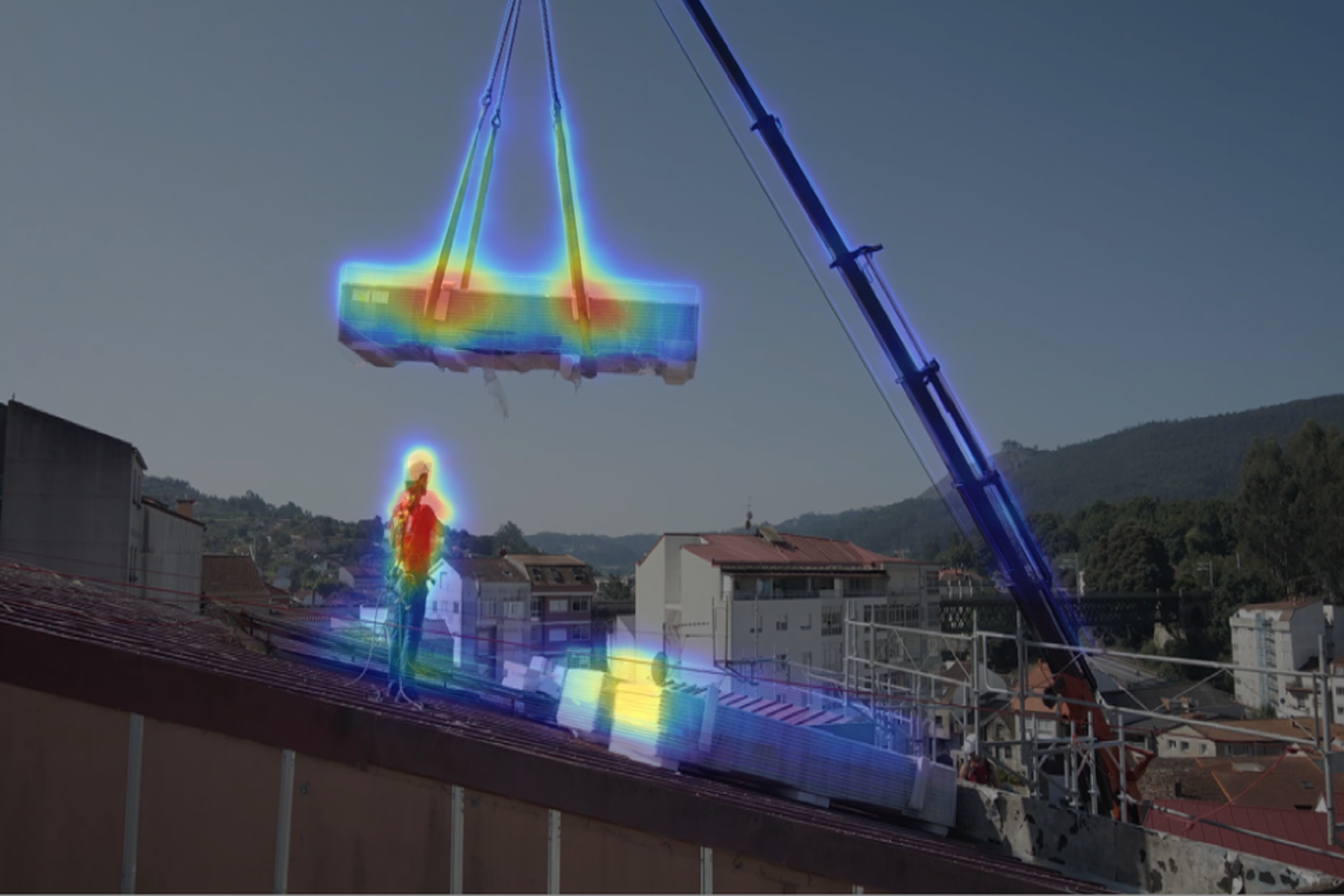} \\
  \includegraphics[width=0.45\textwidth]{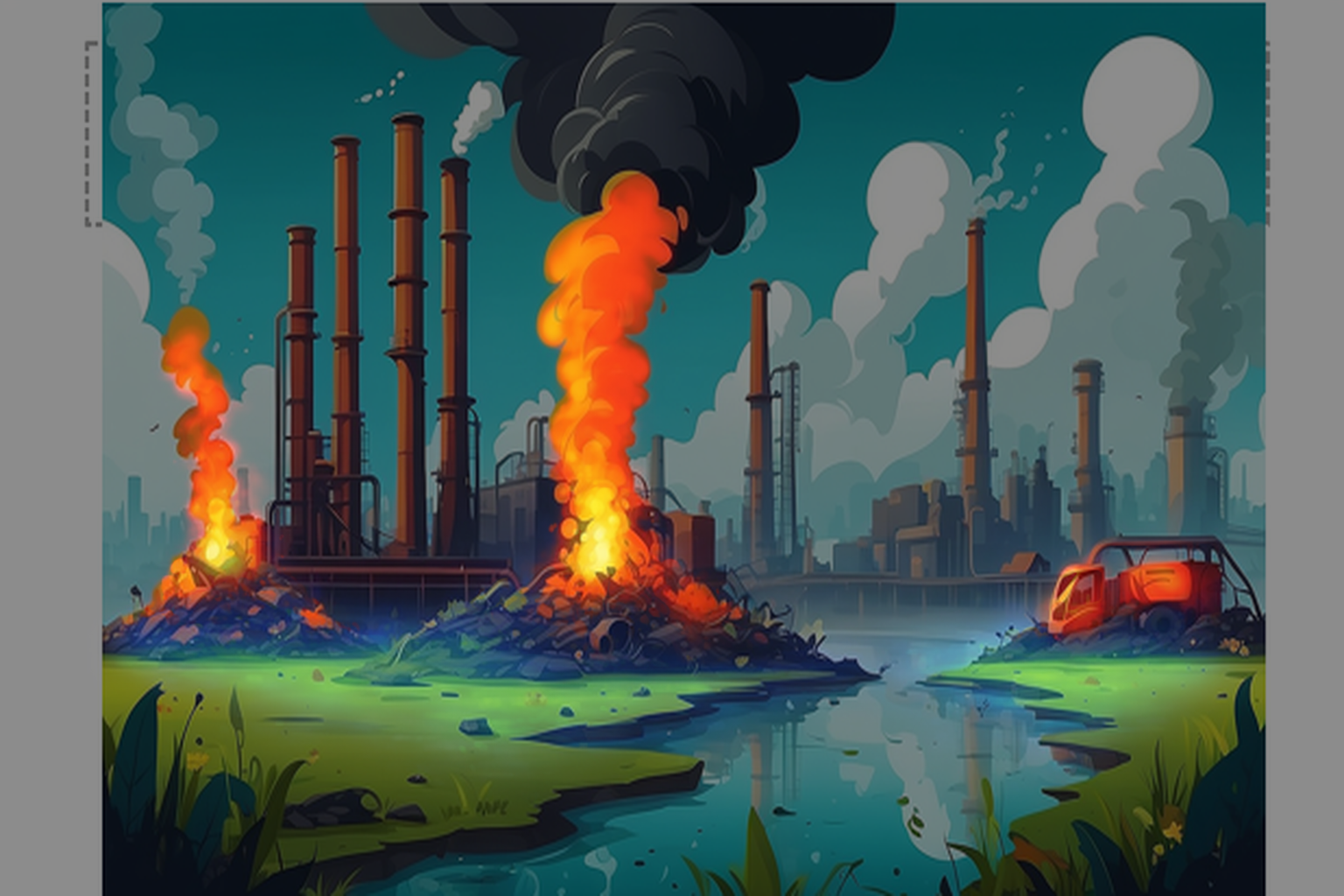} &
  \includegraphics[width=0.45\textwidth]{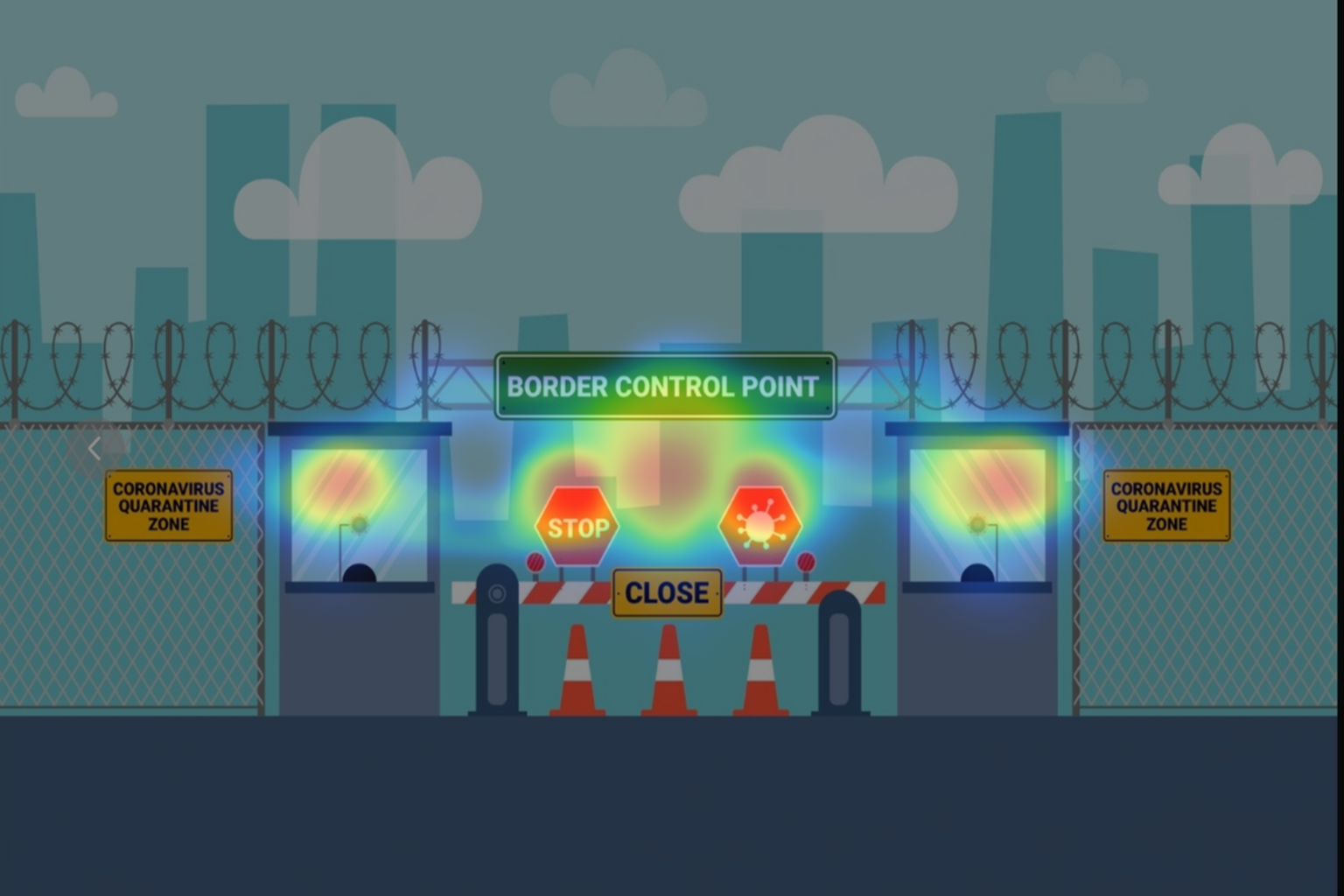} \\[4pt]
  \includegraphics[width=0.38\textwidth]{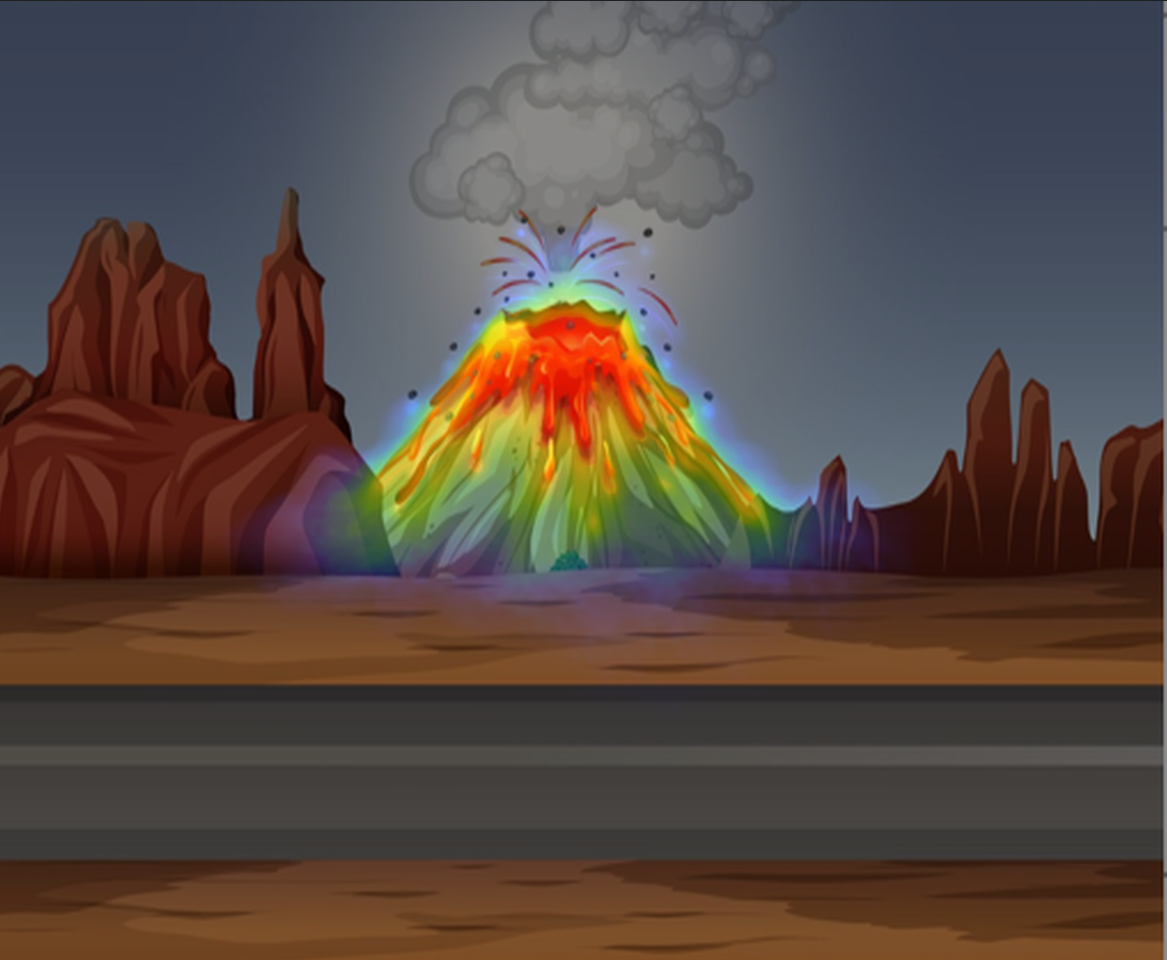} &
  \includegraphics[width=0.45\textwidth]{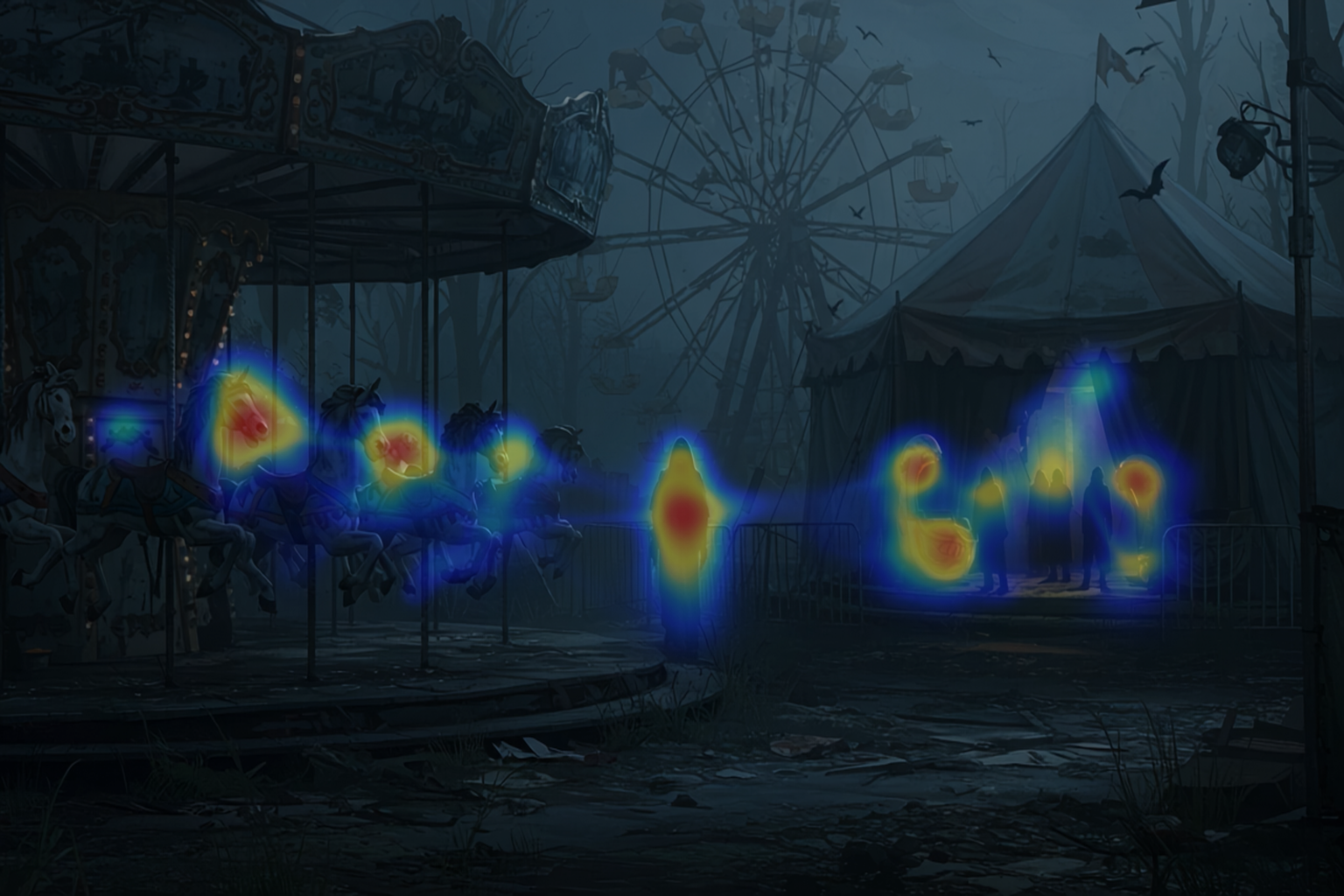} \\[4pt]
\end{tabular}
\caption{GPT-4o saliency heatmaps overlaid on stimuli (images 1--6). Brighter colours indicate higher model saliency.}
\label{fig:ai_heatmap_grid_1}
\end{figure}

\begin{figure}[H]
\centering
\begin{tabular}{cc}
  \includegraphics[width=0.45\textwidth]{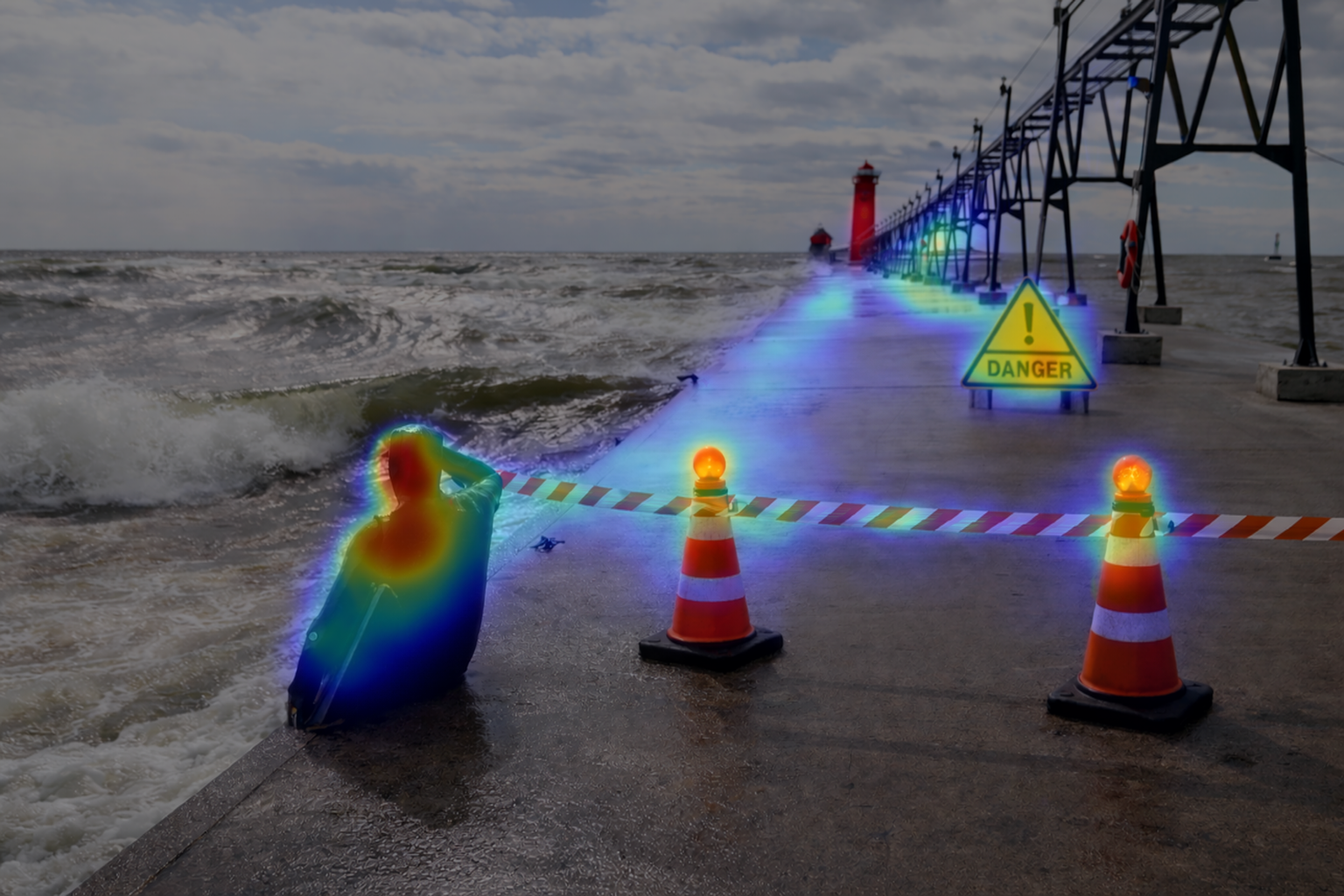} &
  \includegraphics[width=0.45\textwidth]{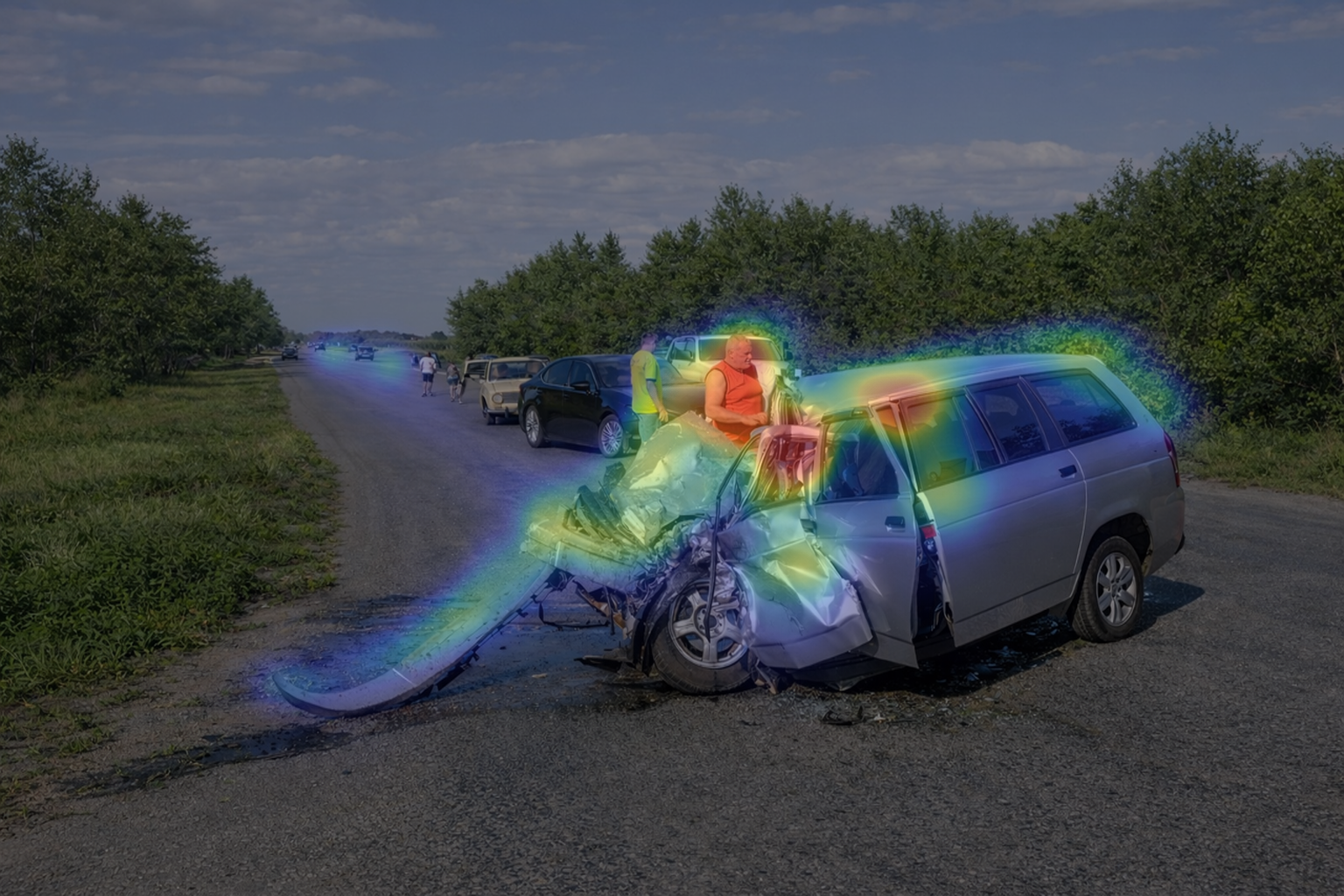} \\[4pt]
  \includegraphics[width=0.45\textwidth]{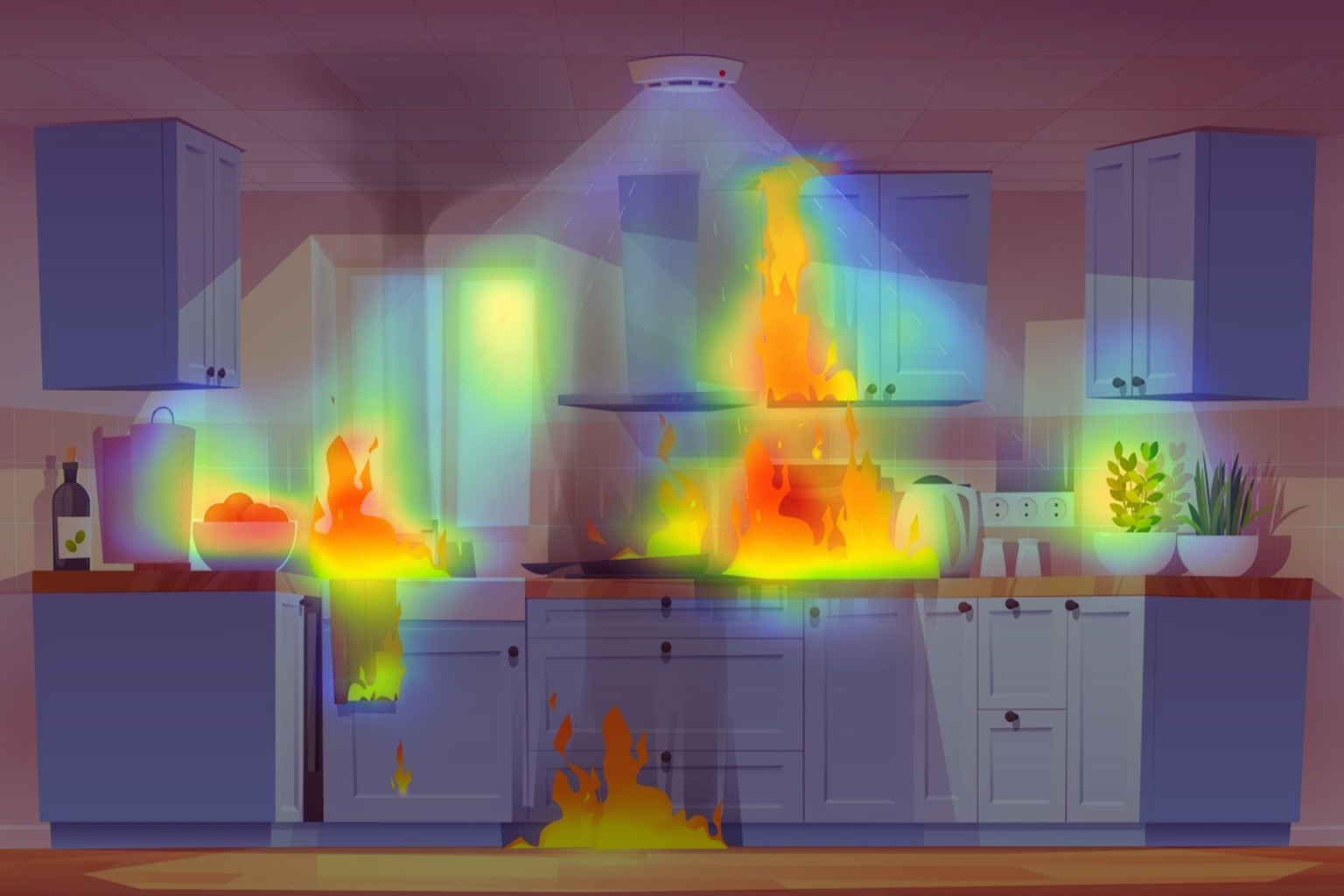} &
  \includegraphics[width=0.45\textwidth]{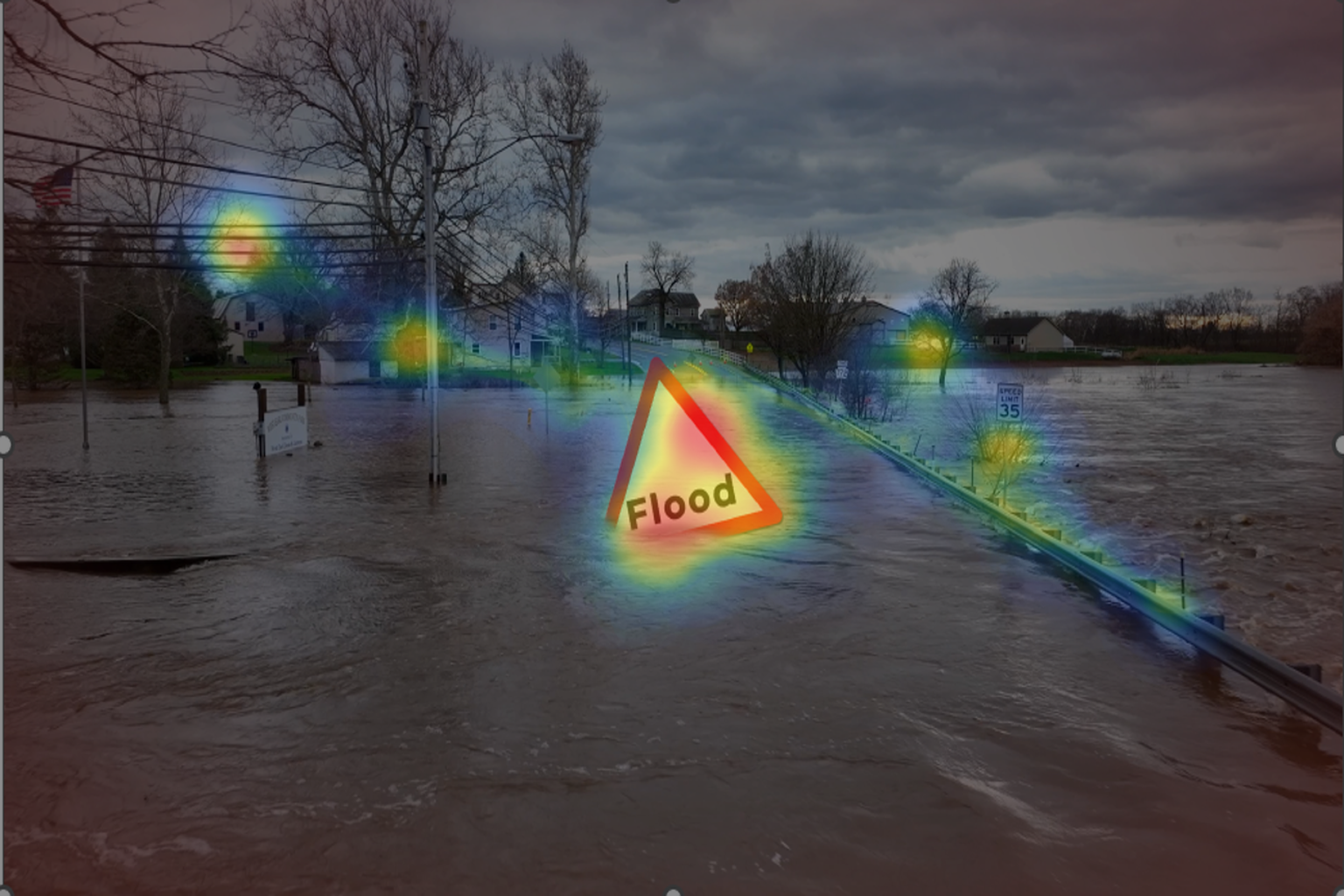} \\
\end{tabular}
\caption{GPT-4o saliency heatmaps overlaid on stimuli (images 7--10). Brighter colours indicate higher model saliency.}
\label{fig:ai_heatmap_grid_2}
\end{figure}

Before interpreting aggregate statistics, Figures~\ref{fig:ai_heatmap_grid_1} and~\ref{fig:ai_heatmap_grid_2} show ten representative stimuli with their corresponding GPT-4o saliency maps overlaid. Brighter colours indicate regions assigned higher saliency by the model. The examples illustrate the range of scene types in the dataset, spanning urban traffic, natural disasters and indoor environments, and give a qualitative sense of where the model concentrates attention across different scene categories before the quantitative metrics are reported below. Aggregate statistics are summarised in Table~\ref{tab:gpt4o_results}.

\begin{table}[H]
\centering
\caption{Summary of saliency metrics comparing population-averaged human gaze heatmaps with GPT-4o risk heatmaps ($n=33$ images). Pearson correlation coefficient ($r$) measures linear spatial agreement between heatmaps, Normalised Scanpath Saliency (NSS) quantifies the correspondence between predicted saliency and human fixations, Kullback--Leibler (KL) divergence measures differences between the predicted and human attention distributions, and Area Under the Receiver Operating Characteristic Curve using the Judd formulation (AUC-Judd) evaluates fixation prediction performance. Higher values indicate better performance for Pearson $r$, NSS, and AUC-Judd, whereas lower values indicate better performance for KL divergence.}
\label{tab:gpt4o_results}
\begin{tabular}{lcccc}
\toprule
 & Pearson $r$ ($\uparrow$) & NSS ($\uparrow$) & KL Divergence ($\downarrow$) & AUC-Judd ($\uparrow$) \\
\midrule
Mean & 0.5153 & 0.9876 & 1.7662 & 0.8064 \\
SD   & 0.1173 & 0.3230 & 0.8442 & 0.0763 \\
Min  & 0.1858 & 0.2817 & 0.5931 & 0.5395 \\
Max  & 0.7792 & 1.7418 & 4.1632 & 0.9222 \\
\bottomrule
\end{tabular}
\end{table}

\subsubsection{Quantitative Performance of GPT-4o}
Across all four metrics, GPT-4o risk heatmaps showed a meaningful level of spatial alignment with population-averaged human gaze heatmaps. The mean AUC-Judd of $0.8064 \pm 0.0763$ substantially exceeds the chance baseline of 0.5, indicating that the AI heatmaps reliably rank fixated locations above non-fixated ones. The mean NSS of $0.9876 \pm 0.3230$ indicates that the AI predictions assign above-average values at the specific pixel locations where human gaze was recorded. The Pearson correlation of $0.5153 \pm 0.1173$ reflects moderate but consistent spatial co-occurrence across the full stimulus set. The mean KL divergence of $1.7662 \pm 0.8442$ reflects variability in distributional overlap across stimuli, which is discussed further below.

\subsubsection{Spatial Alignment Across Scene Images}
\begin{figure}[H]
\centering
\begin{tabular}{cc}
  \includegraphics[width=0.45\textwidth]{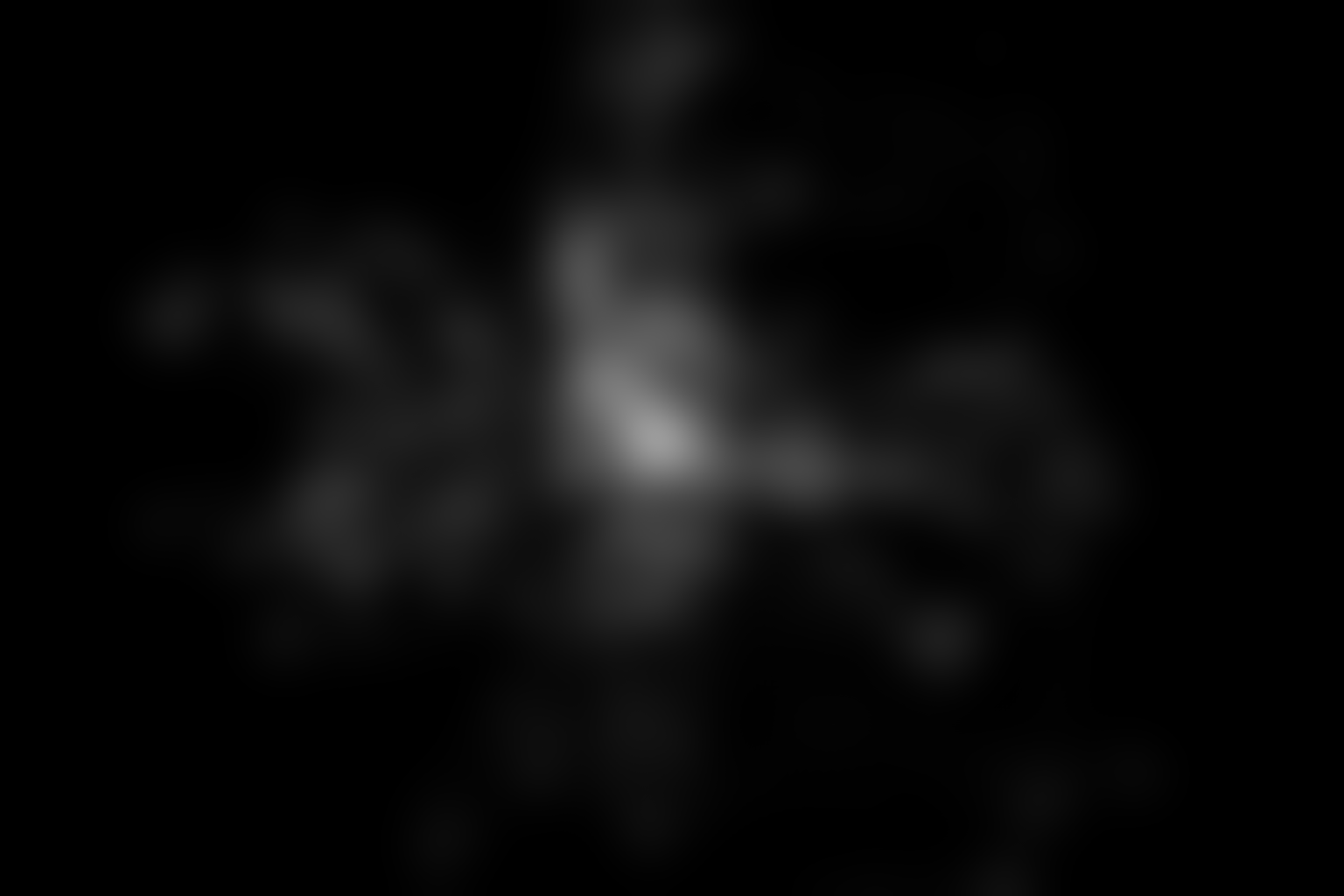} &
  \includegraphics[width=0.45\textwidth]{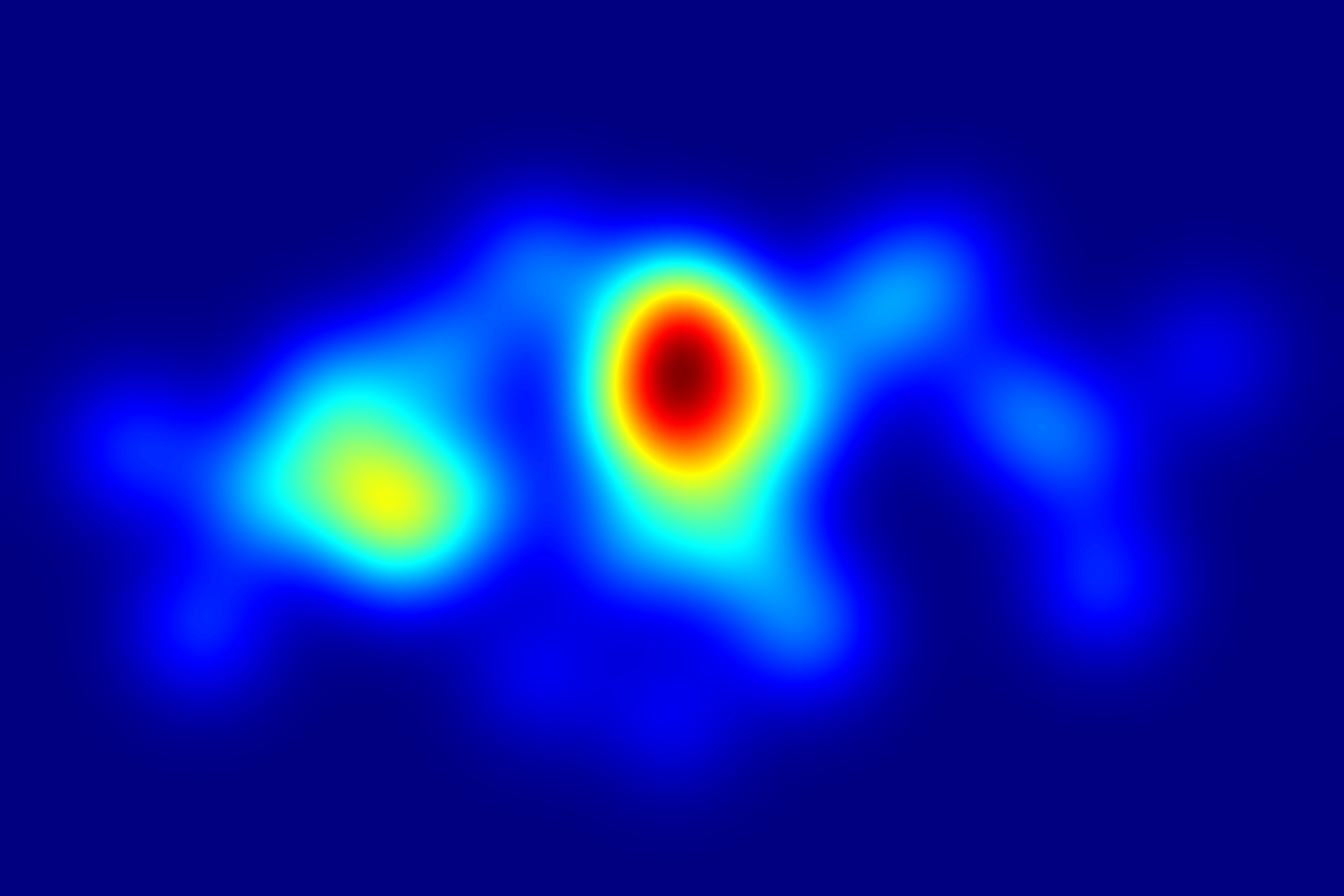} \\
  (a) Human gaze heatmap ($N=10$) & (b) GPT-4o saliency map \\
\end{tabular}
\caption{Qualitative comparison for Image~05: human reference (left) versus AI output (right), $1{,}536\times1{,}024$ pixels.}
\label{fig:qual_05}
\end{figure}

Alignment varied across the stimulus set. Four images achieved the strongest performance across the four metrics: Images~02, 05, 19, and~20. Image~05 obtained the highest Pearson correlation ($0.78$) and the highest Normalised Scanpath Saliency score ($1.74$). Image~02 produced the highest AUC-Judd value ($0.92$) and the lowest KL divergence ($0.59$). Images~19 and~20 also achieved high NSS ($1.50$ and $1.29$ respectively) and AUC-Judd ($0.88$ and $0.91$ respectively). In all four cases, the scene contains visually salient elements associated with potential risk that occupy a specific and well-defined spatial region. This pattern suggests that GPT-4o attains stronger alignment when risk-related content is spatially localised rather than diffuse. Figure~\ref{fig:qual_05} shows Image~05, which is among the stimuli with the strongest agreement between human gaze and the GPT-4o map.

\begin{figure}[H]
\centering
\begin{tabular}{cc}
  \includegraphics[width=0.45\textwidth]{images/avg_win_22.png} &
  \includegraphics[width=0.45\textwidth]{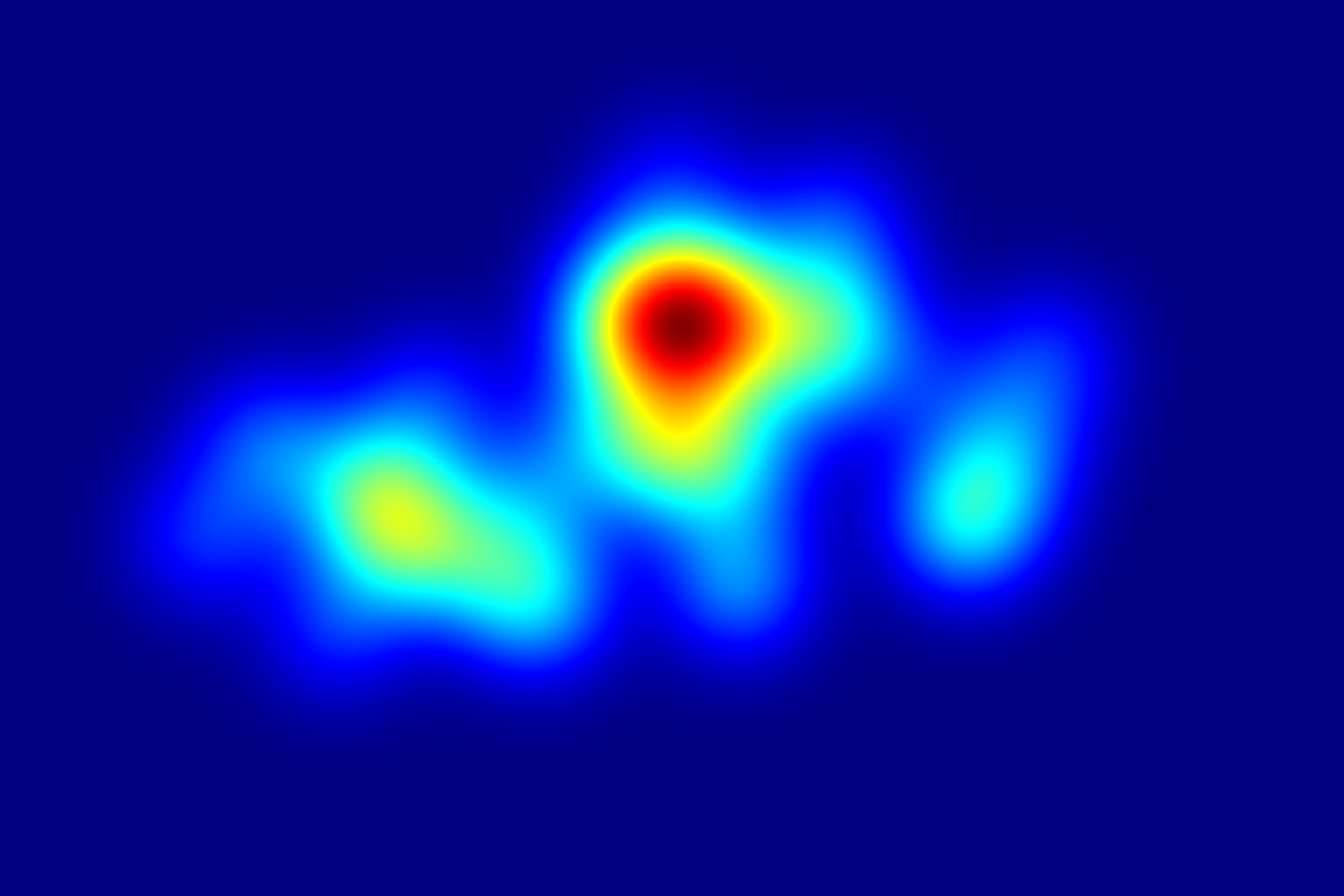} \\
  (a) Human gaze heatmap ($N=10$) & (b) GPT-4o saliency map \\
\end{tabular}
\caption{Qualitative comparison for Image~22: a case with the highest KL divergence in the dataset, consistent with more diffuse spatial overlap between maps.}
\label{fig:qual_22}
\end{figure}

A small number of images showed weaker alignment, most likely due to more diffuse human gaze patterns in those scenes, where attention was spread broadly rather than concentrated on a specific region. These cases contribute to the higher variability observed in Kullback--Leibler divergence. Figure~\ref{fig:qual_22} illustrates Image~22, which has the largest Kullback--Leibler divergence value in the dataset ($4.16$), consistent with more diffuse spatial structure in the maps. No explicit risk-level labels were assigned to the stimuli during the study design, as the primary objective was to evaluate spatial alignment between AI predictions and human gaze rather than to classify scenes by risk level.

\subsubsection{Comparison Across Models}
To assess whether the observed spatial alignment is specific to GPT-4o or generalises across vision-language models, the same pipeline was applied to three additional models: Gemini Pro, Gemini Flash, and Claude. Each model received identical prompts and stimuli; saliency maps were generated and evaluated using the same four metrics.

\begin{table}[H]
\centering
\caption{Comparison of vision-language model saliency maps against population-averaged human gaze heatmaps ($n=33$ images). Results are reported as mean $\pm$ standard deviation for Pearson correlation coefficient ($r$), Normalised Scanpath Saliency (NSS), Kullback--Leibler (KL) divergence, and Area Under the Receiver Operating Characteristic Curve using the Judd formulation (AUC-Judd). Higher values indicate better performance for Pearson $r$, NSS, and AUC-Judd, whereas lower values indicate better performance for KL divergence. Bold values indicate the best-performing model for each metric.}
\label{tab:cross_model}
\small
\resizebox{\linewidth}{!}{%
\begin{tabular}{lcccc}
\toprule
Model & Pearson $r$ ($\uparrow$) & NSS ($\uparrow$) & KL Divergence ($\downarrow$) & AUC-Judd ($\uparrow$) \\
\midrule
GPT-4o       & $0.5153\pm0.1173$ & $0.9876\pm0.3230$ & $\mathbf{1.7662\pm0.8442}$ & $0.8064\pm0.0763$ \\
Gemini Pro   & $\mathbf{0.5706\pm0.1182}$ & $\mathbf{1.1306\pm0.3407}$ & $2.7496\pm1.2765$ & $\mathbf{0.8398\pm0.0635}$ \\
Gemini Flash & $0.5135\pm0.1399$ & $1.0062\pm0.3571$ & $2.0175\pm0.9959$ & $0.8208\pm0.0778$ \\
Claude       & $0.4251\pm0.1565$ & $0.8270\pm0.3693$ & $2.3545\pm1.0848$ & $0.7790\pm0.0848$ \\
\bottomrule
\end{tabular}}
\end{table}

Two findings emerge from Table~\ref{tab:cross_model}. First, all four models exceed the AUC-Judd chance baseline of 0.5 and yield positive NSS scores, confirming that the pipeline generalises across vision-language models without modification. Second, the models differ in alignment quality. Gemini Pro achieves the highest Pearson correlation ($0.5706$), NSS ($1.1306$), and AUC-Judd ($0.8398$) among the four models. However, GPT-4o produces the lowest KL divergence ($1.77$), indicating that its predicted saliency distribution most closely matches the shape of the human attention distribution. Gemini Pro's higher KL divergence ($2.75$) suggests that while it captures spatial location well, its predicted intensity distribution is less aligned with human gaze patterns. Gemini Flash achieves intermediate performance across all four metrics, while Claude produces the weakest alignment overall. These results indicate that the model's capability in structured spatial reasoning influences the degree of correspondence to human visual attention in safety-relevant scenes.

\subsubsection{Relationship to Negativity Bias Theory}
As introduced in Section~\ref{sec:background}, negativity bias theory predicts that both humans and systems trained on human-generated content will preferentially attend to elements representing potential threats \cite{Baumeister2001Bad,Rozin2001Negativity,Ohman2001Snake}. The positive mean NSS score of 0.99 is broadly consistent with this prediction, showing that GPT-4o assigns above-average values at the same locations where human participants directed their gaze. This suggests that GPT-4o implicitly captures something of the human tendency to prioritise potentially risky regions.

\section{Discussion}
\label{sec:discussion}
This study addressed two objectives. First, an eye-tracking dataset was constructed in which ten participants viewed 33 environmental scene images spanning a range of everyday scene types. Individual gaze heatmaps were generated for each participant and combined pixel-wise to produce 33 population-level attention maps representing collective human visual attention. Second, an automated Python pipeline was developed to generate spatially grounded AI saliency heatmaps for the same stimuli using GPT-4o through the OpenAI Vision API, without any training on eye-tracking data. The spatial correspondence between AI predictions and human gaze was then
quantified using four complementary saliency metrics across all 33 stimulus images.

The results demonstrate a meaningful level of spatial alignment between GPT-4o saliency heatmaps and population-averaged human gaze heatmaps. The mean AUC-Judd of $0.8064$, mean NSS of $0.9876$, and Pearson correlation of $0.5153$ collectively indicate that GPT-4o, without any training on eye-tracking data, can identify regions that broadly correspond to where humans direct visual attention in safety-relevant scenes. The observed alignment is consistent with negativity bias theory \cite{Baumeister2001Bad,Rozin2001Negativity,Ohman2001Snake}.

This alignment may reflect GPT-4o's training on large amounts of human-generated content, through which the model may have learned which scene elements humans typically treat as potentially threatening. However, this does not imply that the model experiences negativity bias in a psychological sense. Future work with controlled stimuli containing neutral and risk-relevant versions of the same scene would be needed to confirm this interpretation.

Several limitations of this study should be acknowledged. The AI maps are constructed from a finite set of weighted locations, returned as JSON, and then rasterised as disks before Gaussian smoothing; this sparse parameterisation can miss fine-grained spatial detail present in dense human gaze distributions. In addition, the AI-generated heatmaps depend on the specific prompt design used to query the vision-language model, and different prompting strategies may produce variations in spatial predictions. API sampling variability may also contribute to run-to-run differences when responses are regenerated. Furthermore, the participant sample of ten university students is relatively small, which may limit the generalisability of the population-averaged gaze heatmaps.

The comparison across models further demonstrated that the pipeline is not specific to GPT-4o: all four models exceeded the AUC-Judd chance baseline. Gemini Pro achieved the strongest spatial localisation on three of four metrics, while GPT-4o produced the closest distributional match to human attention as measured by KL divergence.

\section{Conclusions}
\label{sec:conclusions}
The spatial correspondence between AI-generated risk heatmaps and human gaze heatmaps, quantified using four complementary saliency metrics across all 33 stimulus images, demonstrates that GPT-4o can identify regions of potential risk that broadly correspond to where humans naturally direct visual attention in safety-relevant environments. These findings suggest that large vision-language models can support safety-aware AI applications that require an understanding of human visual attention, without requiring any eye-tracking training data.

Future work could address the limitations identified in this study by recruiting a larger, more diverse participant sample, incorporating expert-rated risk-level labels to enable subgroup analyses and statistical testing of differences in alignment across scene types, and exploring finer-grained spatial prompting strategies or alternative vision-language models. Extending the paradigm to dynamic video stimuli would also allow for a more ecologically valid assessment of how AI risk predictions track human attention over time. From a robotics perspective, a natural next step would be to integrate vision-language model saliency predictions into a robot perception stack, enabling onboard hazard detection without dedicated eye-tracking hardware. Evaluating whether such predictions can guide real-time path planning or trigger safety responses in a mobile robot platform would directly test the practical value of the spatial alignment demonstrated here.

Overall, this study demonstrates that large vision-language models such as GPT-4o can produce spatially grounded risk predictions that meaningfully correspond to human visual attention across diverse everyday scene types, without requiring any eye-tracking training data. These findings provide a foundation for developing safety-aware AI systems that better align with human perception of potential risks in real-world environments. This may support the design of robots that can operate safely alongside humans without requiring dedicated gaze-tracking hardware.

\appendix

\section{Final GPT-4o Prompt (Verbatim)}
\label{app:prompt}

\begin{verbatim}
You are a visual-saliency expert predicting a population-averaged 
fixation density map.

The image will be viewed freely for 10 seconds by many observers. 
Your output will be rasterised as weighted disks and Gaussian-blurred 
to form a smooth density map -- so predict the SPATIAL DISTRIBUTION 
of attention across the whole scene, not just isolated objects.

Return ONLY a JSON object, no markdown:
{"fixations": [
  {"x": 0.52, "y": 0.38, "w": 1.0},
  {"x": 0.28, "y": 0.55, "w": 0.7}
]}

Rules:
- x: 0.0 (left) to 1.0 (right); y: 0.0 (top) to 1.0 (bottom)
- w: saliency weight 0.1-1.0
- Provide 20-25 fixation points to cover the full spatial distribution
- Priority regions (w 0.7-1.0): faces, people, vehicles, hazards,
  animals, readable text/signs
- Secondary regions (w 0.3-0.6): scene boundaries, supporting
  context, secondary objects
- Background (w 0.1-0.2): empty sky, blank walls, plain road surface
- Apply mild centre bias: central area slightly favoured when no
  strong peripheral content exists
- Spread points across ALL meaningful areas -- avoid clustering
- ONLY the JSON object, nothing else
\end{verbatim}

\section{Full Gaze CSV Column Reference}
\label{app:gaze_csv}

The exported gaze CSV contains the following columns. Section and recording identifiers are stored as full UUID strings in the raw export; abbreviated forms (e.g.\ \texttt{d8**12}, \texttt{7850**}) are used in Table~\ref{tab:gaze_csv_excerpt} for readability.

\begin{table}[H]
\centering
\caption{Complete column listing of the exported gaze CSV.\label{tab:gaze_csv_excerpt}}
\begin{tabularx}{\textwidth}{llX}
\toprule
\textbf{Column} & \textbf{Example value} & \textbf{Description} \\
\midrule
\texttt{section id}         & d8004db2-78504cdd-\ldots & UUID identifying the recording section \\
\texttt{recording}          & 78504cdd-\ldots           & UUID of the parent recording session \\
\texttt{timestamp {[}ns{]}} & 1.77213E+18               & Gaze sample timestamp in nanoseconds \\
\texttt{gaze x {[}px{]}}    & 742.325                   & Horizontal gaze coordinate (scene camera pixels) \\
\texttt{gaze y {[}px{]}}    & 374.311                   & Vertical gaze coordinate (scene camera pixels) \\
\bottomrule
\end{tabularx}
\end{table}


\end{document}